\documentclass[12pt]{spieman}  

\usepackage{amsmath,amsfonts,amssymb}
\usepackage{graphicx}
\usepackage{setspace}
\usepackage{tocloft}
\usepackage{subcaption}
\usepackage{booktabs}
\usepackage{multirow}
\usepackage{array}
\definecolor{darkgreen}{RGB}{0,128,0}
\definecolor{darkred}{RGB}{180,0,0}
\usepackage[table]{xcolor}

\title{Fourier Transform Multiple Instance Learning for Whole Slide Image Classification}

\author[a]{Anthony Bilic}
\author[a]{Guangyu Sun}
\author[a]{Ming Li}
\author[c]{Md Sanzid Bin Hossain}
\author[a]{Yu Tian}
\author[b]{Wei Zhang}
\author[a, c]{Laura Brattain}
\author[c]{Dexter Hadley}
\author[a]{Chen Chen}
\affil[a]{Institute of Artificial Intelligence (IAI), 4000 Central Florida Blvd, Orlando, FL 32816, USA}
\affil[b]{Department of Computer Science, University of Central Florida, 4000 Central Florida Blvd, Orlando, FL 32816, USA}
\affil[c]{College of Medicine, University of Central Florida, 6850 Lake Nona Blvd, Orlando, FL 32827, USA}

\cftpagenumbersoff{figure}
\cftpagenumbersoff{table} 
\begin{document} 
\maketitle

\begin{abstract}
\\ \\
\noindent\textbf{Purpose:}\hspace*{0.5em} Whole slide image (WSI) classification relies on Multiple Instance Learning (MIL) with spatial patch features, but current methods struggle to capture global dependencies due to the immense size of WSIs and the local nature of patch embeddings. This limitation hinders the modeling of coarse structures essential for robust diagnostic prediction.
\\ \\
\noindent\textbf{Approach:}\hspace*{0.5em}We propose Fourier Transform Multiple Instance Learning (FFT-MIL), a framework that augments MIL with a frequency-domain branch to provide compact global context. Low-frequency crops are extracted from WSIs via the Fast Fourier Transform and processed through a modular FFT-Block composed of convolutional layers and Min-Max normalization to mitigate the high variance of frequency data. The learned global frequency feature is fused with spatial patch features through lightweight integration strategies, enabling compatibility with diverse MIL architectures.
\\ \\ 
\noindent\textbf{Results:}\hspace*{0.5em}FFT-MIL was evaluated across six state-of-the-art MIL methods on three public datasets (BRACS, LUAD, and IMP). Integration of the FFT-Block improved macro F1 scores by an average of 3.51\% and AUC by 1.51\%, demonstrating consistent gains across architectures and datasets.
\\ \\
\noindent\textbf{Conclusions:}\hspace*{0.5em}FFT-MIL establishes frequency-domain learning as an effective and efficient mechanism for capturing global dependencies in WSI classification, complementing spatial features and advancing the scalability and accuracy of MIL-based computational pathology. \textit{Code publicly available at  \href{https://github.com/irulenot/FFT-MIL}{https://github.com/irulenot/FFT-MIL}.}
\end{abstract}


{\noindent \footnotesize\textbf{*}Anthony Bilic, \linkable{an609701@ucf.edu} }

\begin{spacing}{2}   

\section{Introduction}
\label{sec:intro}
Computational pathology has transformed clinical diagnostics by efficiently digitizing haematoxylin and eosin (H\&E)-stained whole slide images (WSIs) using automated digital scanners~\cite{hosseini2024computational}. This innovation has spurred a surge in artificial intelligence (AI) research, with the potential to automate clinical diagnosis, predict patient prognosis, and therapeutic response~\cite{song2023artificial}. However, due to the enormous size of each WSI, often exceeding 100 million pixels, applying AI to WSIs faces two significant challenges. First, annotating WSIs requires substantial time from pathologists due to the extensive area of these images. Second, current deep learning approaches cannot process an entire WSI directly due to hardware constraints~\cite{vorontsov2023virchow}.

To alleviate the cost of acquiring comprehensive pixel-level annotations, many current methods instead use slide-level annotations, which assign a single label to each WSI and are easier to obtain~\cite{hosseini2024computational}. Using slide-level annotations, Multiple Instance Learning (MIL)~\cite{ilse2018attention} has become the most widely used framework in computational pathology~\cite{tang2024feature}. MIL partially relaxes the limitations of performing tasks on WSIs with its weakly supervised approach by using unlabeled WSI patches for downstream analysis~\cite{gadermayr2024multiple, li2023task}. The MIL framework pipeline can be abstracted into four main stages: First, either all or a subset of patches are selected from the WSIs for analysis. Second, the selected patches are converted into patch features, typically using a pretrained natural image model such as the ResNet50~\cite{he2016deep} model trained on the ImageNet dataset~\cite{deng2009imagenet}. Third, these patch features are aggregated to form a combined structured feature representation. Finally, a collective processing stage assigns a label to the entire WSI~\cite{gadermayr2024multiple}.

Although MIL has achieved strong performance in WSI classification, it struggles to effectively capture long-range dependencies~\cite{deng2024cross, yang2024scmil}. This limitation is critical because WSIs contain both fine-grained cellular details and coarse-grained structures such as cancer-associated stroma and epithelial tissue~\cite{zhang2024prompting}. A common strategy to address this challenge is multi-magnification analysis~\cite{liu2024wsi, li2021dual}, which enhances global context modeling by combining information across multiple resolutions and linking fine-grained patch details with broader structural context. In contrast, we propose leveraging the Fast Fourier Transform (FFT) to obtain a single, compact, and information-rich representation of the entire WSI, providing an alternative mechanism for capturing global context.

In deep learning, frequency analysis is typically applied within architectures as an auxiliary operation on spatial features, extending the modeling capacity of Convolutional Neural Networks (CNNs) and Transformers. Unlike prior approaches that apply frequency analysis only as an augmentation to spatial features, we introduce a separate branch that directly processes frequency-domain inputs to learn global representations, which are then fused with spatial features for downstream tasks. A key insight enabling this design comes from image compression literature~\cite{pandey2015image}, which shows that most of the signal energy in frequency-transformed images is concentrated in the low-frequency components. Leveraging this property, we derive a compact low-frequency crop, substantially smaller than the original WSI, that preserves global information and enables efficient modeling of long-range dependencies.

Learning directly from the frequency domain poses a significant challenge due to the high variance inherent in frequency data~\cite{chu2023rethinking}. Prior works address this by designing specialized architectures~\cite{li2020fourier, chi2020fast, rippel2015spectral} that rely on the Inverse Fast Fourier Transform (iFFT) to project frequency features back into the spatial domain for fusion. In contrast, we propose a frequency feature normalization scheme that encodes the frequency input with convolutional layers followed by Min-Max normalization. Min-Max normalization is particularly suitable as it avoids reliance on standard deviation and has demonstrated success in approximating non-linear functions in homomorphic encryption~\cite{aboulatta2019stabilizing}. This choice mitigates the high variance of frequency data, maps features into a consistent space, and enables stable fusion with spatial representations.

We propose Fourier Transform Multiple Instance Learning (FFT-MIL), a framework for WSI classification that leverages frequency-domain information to enhance global context modeling. Our contributions are threefold: (1) We design a preprocessing pipeline that extracts low-frequency crops from WSIs, producing compact and information-dense representations that capture slide-level dependencies. (2) We introduce the FFT-Block, a modular component that learns directly from frequency-domain inputs using convolutional layers followed by Min-Max normalization, enabling effective fusion of frequency-derived global features with spatial representations. (3) We demonstrate that FFT-MIL consistently improves performance when integrated with six state-of-the-art MIL architectures across three public datasets, increasing average F1 scores by 3.51\% and AUC by 1.51\%. These results establish frequency-domain learning as an effective means of augmenting spatial models for improved long-range dependency modeling in WSIs.

\section{Related Works}
\subsection{Multiple Instance Learning}
The primary constraint in whole slide image (WSI) classification is effectively modeling the large number of patches required to process large resolution images~\cite{li2023task}. Earlier MIL-based approaches, patches are encoded using a natural image encoder followed by global pooling or self-attention~\cite{ilse2018attention, lu2021data}, but several limitations persist. First, spatial relationships between patches are weakly modeled. To address this, recent methods incorporate graph neural networks~\cite{shi2024integrative}, multi-scale architectures~\cite{liu2024wsi}, and patch coordinate pairs~\cite{yang2024scmil} to capture inter-patch relationships. Second, global contextual information is often underutilized, as patch-level features alone fail to capture coarse-grained patterns such as tumor-stroma interactions. This has motivated the use of hierarchical architectures that use multiple magnifications to better capture global dependencies~\cite{deng2024cross, liu2024wsi}. Third, the imbalance of positive and negative instances in bags introduces redundancy and interferes with attention mechanisms. Methods such as patch clustering and global feature aggregation have been proposed to mitigate this issue and enhance instance diversity~\cite{zhang2024prompting, lin2023interventional, zhu2024dgr, zhang2024attention}. Fourth, the quadratic complexity of self-attention makes it infeasible for WSIs with tens of thousands of patches, leading to the application of linear approximations, low-rank attention, and retention-based mechanisms~\cite{yang2024scmil, xiang2023exploring, chu2024retmil}. Finally, to manage the overwhelming number of patches, sampling and feature reduction techniques are employed~\cite{neto2024interpretable, li2023task}. However, due to sampling often discarding spatial context, some works~\cite{zhang2024ihcsurv} propose more sophisticated sampling approaches such as region-aware clustering. 

We address the challenge of modeling global dependencies by proposing an alternative to hierarchical architectures that use multi-resolution spatial inputs from downsampled image pyramids. Our parallel and modular design incorporates global context into existing MIL frameworks through a single, compact, and information-rich frequency representation of WSIs.

\subsection{Frequency Architectures}
Current methods integrate frequency analysis by applying the Fourier Transform to spatial features within specialized architectures. In transformers, this improves modeling of high-frequency details~\cite{wang2024freqformer, xu2024dual, pathak2022fourcastnet}, while in CNNs it enhances access to low-frequency information, mitigating the constraint of local receptive fields~\cite{paing2023adenoma, zhang2022swinfir, suvorov2022resolution}. Furthermore, several studies report that frequency-domain representations capture structural information that is difficult to model purely in the spatial domain~\cite{song2023fourier, huang2022deep}.

Unlike existing methods, our approach directly processes frequency-domain representations of images rather than intermediate spatial features. While prior frequency-based architectures rely on the iFFT to project frequency features back to the spatial domain before fusion~\cite{li2020fourier, suvorov2022resolution, nair2020fast, chu2023rethinking, huang2022deep, zhang2022swinfir, wen2022u, pathak2022fourcastnet, paing2023adenoma, zheng2024fouriermil}, we instead apply Min-Max normalization, enabling direct fusion of frequency and spatial features.

\section{Methodology}
\label{sec:method}
\begin{figure}[H]
\includegraphics[width=\textwidth]{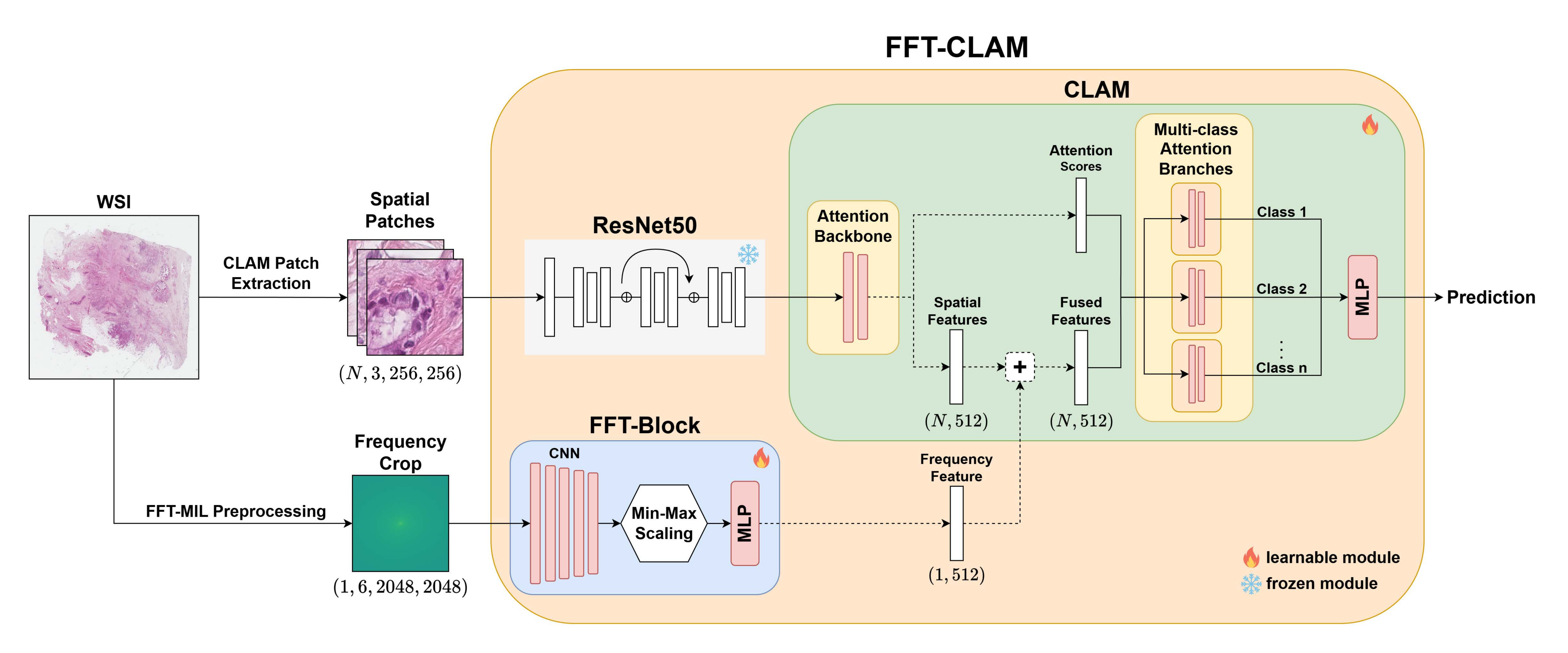}
\caption{Overview of the proposed Fourier Transform Multiple Instance Learning (FFT-MIL) framework integrated with CLAM~\cite{lu2021data} for WSI classification. The FFT-Block extracts a global frequency feature from a given WSI, which is fused with the output of CLAM’s~\cite{lu2021data} attention backbone via addition to introduce global context at a stage where patch-level information has been aggregated. While illustrated with CLAM~\cite{lu2021data}, the FFT-Block is modular and can be integrated into other MIL methods in a similar fashion.}
\label{fig:diagram}
\end{figure}

The proposed Fourier Transform Multiple Instance Learning (FFT-MIL) framework augments existing MIL methods with a frequency-domain branch to improve global context modeling in WSI classification. Figure~\ref{fig:diagram} shows its integration into CLAM~\cite{lu2021data}, which we select as the primary baseline due to its strong performance and widespread adoption in the MIL literature. To demonstrate the generality of FFT-MIL, we further extend this integration strategy to five additional state-of-the-art MIL frameworks, as detailed in Section~\ref{subsec:comparison}.

FFT-MIL proposes two key additions to MIL-based architectures. First, in Section~\ref{subsec:preprocessing}, we present our preprocessing pipeline for obtaining low-frequency representations of WSIs. Second, in Section~\ref{subsec:block}, we introduce the Fast Fourier Transform Block (FFT-Block), a modular component that uses these representations to inject learned global dependencies into MIL-based models. In addition, Section~\ref{subsec:complexity} provides a comparative complexity analysis of conventional patch processing compared to our proposed frequency preprocessing.

\subsection{Low-Frequency Representation Preprocessing}
\label{subsec:preprocessing}
Patch-wise processing produces an extremely large number of instances, making end-to-end learning computationally infeasible~\cite{vorontsov2023virchow} and limiting the ability to model global dependencies. To address this, we propose learning from a compressed frequency-domain representation that captures long-range context and can be trained end-to-end, which is subsequently fused with MIL architectures for fine-grained analysis.

\begin{figure}[H]
\includegraphics[width=\textwidth]{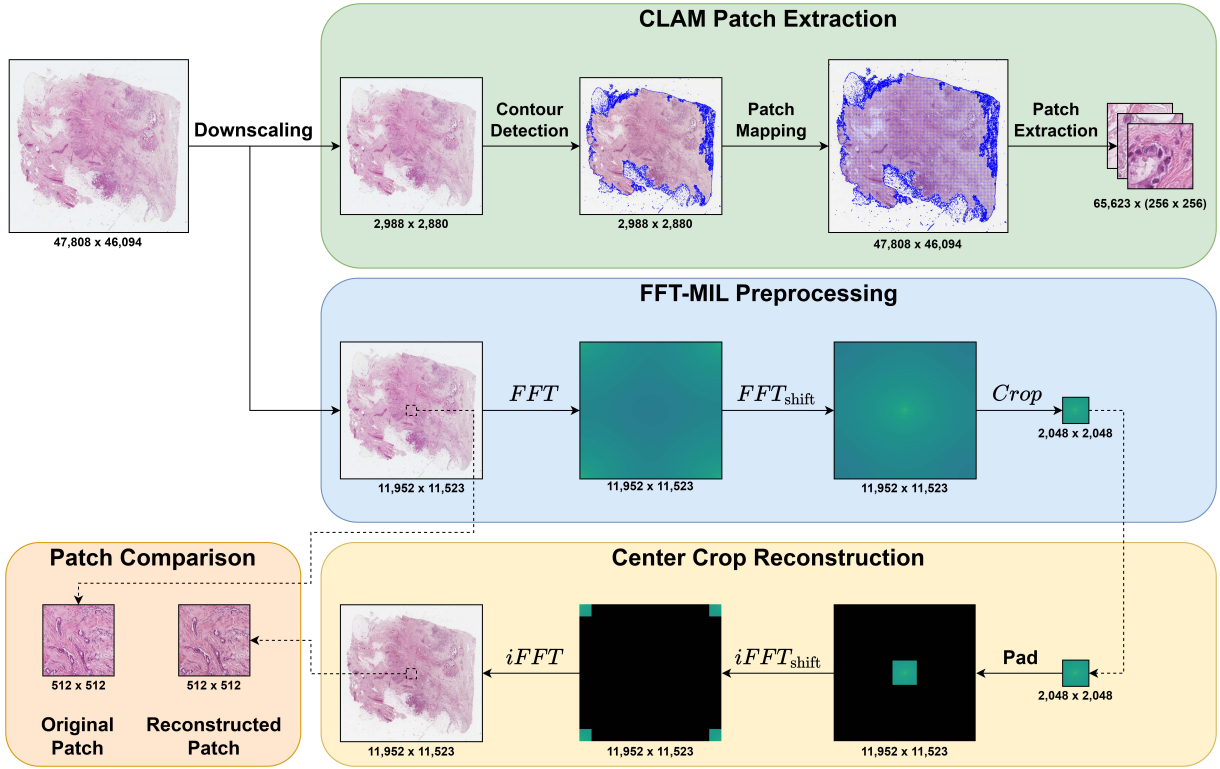}
\caption{Overview of our proposed preprocessing pipeline for obtaining low-frequency representations of WSIs. The CLAM~\cite{lu2021data} patch extraction branch (top) uses a \( 16\times \) to \( 64\times \) downsampled WSI for tissue segmentation, which is then aligned to the full-resolution image for patch extraction. The FFT-MIL branch (middle) operates on a \( 4\times \) downsampled WSI, applying FFT, frequency shift, and center cropping to retain low-frequency components. The reconstruction branch (bottom right), included for visualization purposes only, performs inverse FFT and padding to approximate the original image. A visual comparison of original and reconstructed patches is shown (bottom left).}
\label{fig:preprocessing}
\end{figure}

Figure~\ref{fig:preprocessing} illustrates our pipeline for extracting low-frequency representations of WSIs. Following prior work on natural image statistics, we assume that WSIs consist of independent constant-intensity regions whose sizes follow a power-law distribution~\cite{balboa2001occlusions}. As a result, applying the FFT and zero-frequency centering ($\text{FFT}_{\text{shift}}$) concentrates most of the spectral power at low spatial frequencies, primarily centered and along horizontal and vertical orientations~\cite{ruderman1993statistics}. We exploit this property by extracting a center crop of the frequency image, which retains the majority of slide-level information while substantially reducing the input size for downstream processing. This procedure effectively implements a low-pass filter, suppressing high-frequency noise and preserving global structure~\cite{makandar2015image}.

Our proposed Low-Frequency Representation Preprocessing consists of four steps on a given $4\times$ downscaled WSI. Downscaling is applied to WSIs before frequency preprocessing due to the \( O(N \log N) \) complexity of the FFT~\cite{li2020fourier}, where \( N \) is the number of pixels, making full-resolution processing computationally prohibitive. \textbf{First}, we convert it into a frequency-domain representation.
\begin{equation}
F =
\begin{bmatrix}
\mathcal{FFT}(I_R) \\
\mathcal{FFT}(I_G) \\
\mathcal{FFT}(I_B)
\end{bmatrix}, \quad
\mathcal{FFT}(I_C)(u,v) = \sum_{x=0}^{M-1} \sum_{y=0}^{N-1} I_C(x,y) \cdot e^{-j 2\pi \left( \frac{ux}{M} + \frac{vy}{N} \right)}
\label{eq:fft_representation}
\end{equation}
Here the variable \( I_C(x, y) \) represents the intensity of the color channel \( C \in \{R, G, B\} \) at spatial coordinates \( (x, y) \). The coordinates \( (x, y) \) correspond to the spatial domain, while \( (u, v) \) represent the frequency domain coordinates in the Fourier-transformed space. \( M \) and \( N \) denote the width and height of the image, respectively.

\textbf{Second}, we apply zero-frequency centering to the frequency image representation.
\begin{equation}
F_{\text{shifted}} = \mathcal{FFT}_{\text{shift}}(F) = (-1)^{u+v} \cdot F(u,v)
\end{equation}
Here, the variable \( F(u,v) \) represents the Fourier-transformed image at the frequency domain coordinates \( (u, v) \).

\textbf{Third}, after being centered, we take a \( 2{,}048 \times 2{,}048 \) center crop of the frequency representation. This size is empirically selected based on the trend observed in Figure~\ref{fig:spectra_region}, where larger crop sizes consistently improve performance, as they retain a greater portion of the frequency domain. If the WSI is smaller than \( 2{,}048 \times 2{,}048 \), padding is applied.
\begin{equation}
F_{\text{crop}} = \text{Crop}(F_{\text{shifted}}) =
\begin{cases} 
    F_{\text{shifted}}(u,v), & \quad \text{if } 
    \begin{aligned}
        &\frac{M}{2} - 1024 \leq u < \frac{M}{2} + 1024, \\
        &\frac{N}{2} - 1024 \leq v < \frac{N}{2} + 1024
    \end{aligned} \\
    0, & \quad \text{otherwise}
\end{cases}
\end{equation}
Here, \( M \) and \( N \) are the image dimensions in the frequency domain, representing the number of rows and columns.

The resulting frequency crop is in the form of an imaginary number, which can be represented by magnitude and phase components~\cite{kakarala2012interpreting}. \textbf{Fourth}, we extract these components for two reasons. First, the magnitude and phase are real numbers, which allow us to design the FFT-Block using conventional neural networks, which are widely supported by deep learning libraries. Second, an analysis of directly using frequency data with neural networks finds that activation functions, such as ReLU, will cause many of the negative values to become zero due to data's property of having extremely high variance~\cite{chu2023rethinking}. Using the magnitude, which contains only positive values, we can circumvent this issue. Unlike the magnitude, which is non-negative and unbounded, the phase component ranges between \( [-\pi, \pi] \) and are used directly.
\begin{equation}
\label{eq:magnitude}
M = \text{Magnitude}(F_{\text{crop}})(u,v) = \sqrt{\Re\left(F_{\text{crop}}(u,v)\right)^2 + \Im\left(F_{\text{crop}}(u,v)\right)^2}
\end{equation}
\begin{equation}
\label{eq:phase}
P = \text{Phase}(F_{\text{crop}})(u,v) = \tan^{-1}\left( \frac{\Im\left(F_{\text{crop}}(u,v)\right)}{\Re\left(F_{\text{crop}}(u,v)\right)} \right)
\end{equation}
Here, \( \Re \) and \( \Im \) represent the real and imaginary parts of \( F_{\text{crop}}(u,v) \), respectively, where \( M \in \mathbb{R}_{\geq 0} \) and \( P \in [-\pi, \pi] \). We finally concatenate the magnitude and phase components for processing, which is denoted as $F_{\text{wsi}}$.

\subsection{Fast Fourier Transform Block}
\label{subsec:block}
Previous frequency-based architectures~\cite{li2020fourier, suvorov2022resolution, nair2020fast, chu2023rethinking, huang2022deep, zhang2022swinfir, wen2022u, pathak2022fourcastnet, paing2023adenoma, zheng2024fouriermil} do not apply neural networks directly to frequency inputs, but instead perform frequency analysis on spatial features. Processing frequency data, especially from large resolution images, is challenging due to its dynamic range spanning seven to eight orders of magnitude, in contrast to spatial inputs that are typically normalized to $[0, 255]$ or $[0, 1]$~\cite{chu2023rethinking}. Consequently, prior works apply the iFFT to frequency features before fusion with spatial features. Effective normalization strategies for frequency-domain learning remain an open research problem~\cite{cakaj2023spectral}.

\begin{figure}[H]
\centering
\includegraphics[width=\textwidth]{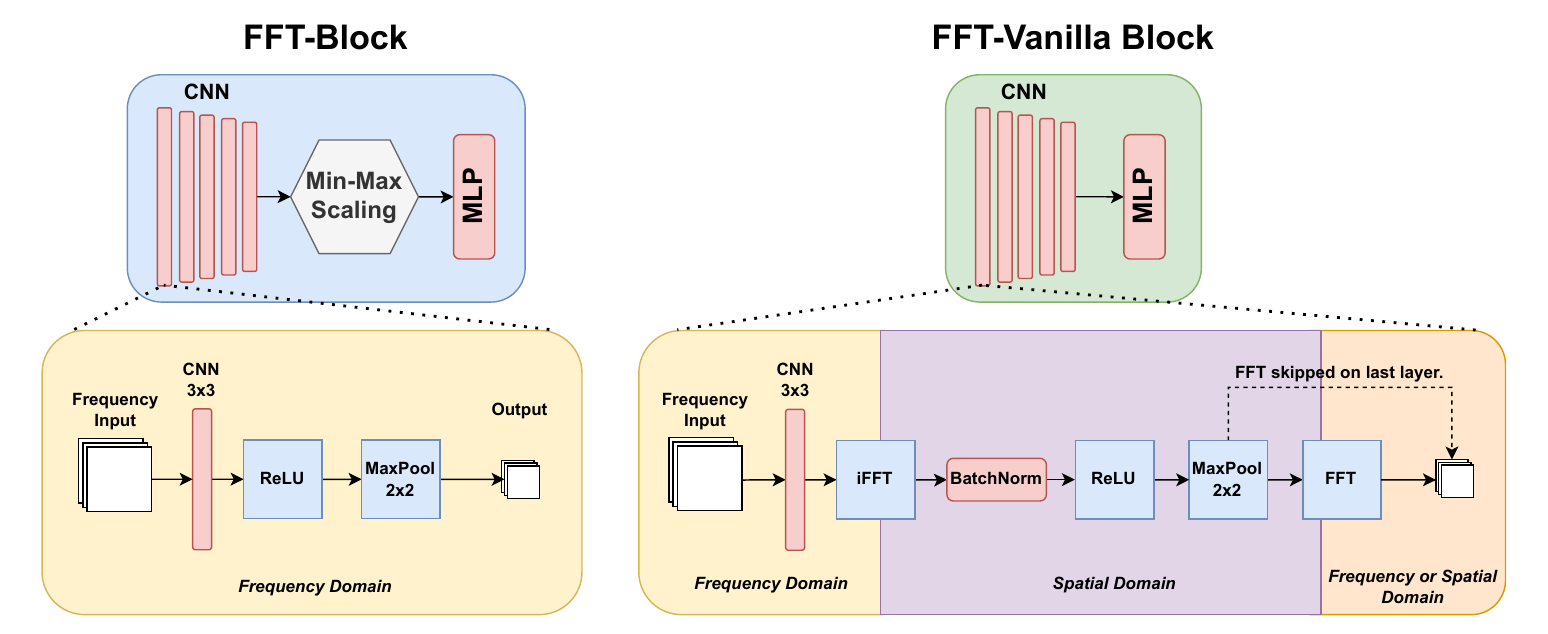}
\caption{
Architectures of our proposed FFT-Block and the FFT-Vanilla Block. \textbf{FFT-Block}: A modular component that operates entirely in the frequency domain using repeated 2D \( 3 \times 3 \) convolutions, ReLU activations, and \( 2 \times 2 \) max pooling. The 2D output is normalized via Min-Max scaling and passed to a multi-layer perceptron  block, producing a global frequency feature for integration with MIL-based architectures or direct classification. \textbf{FFT-Vanilla Block}: A baseline component used to illustrate the role of the iFFT in current frequency-domain architectures. It applies repeated 2D \( 3 \times 3 \) convolutions, each followed by an inverse FFT, Batch Normalization, ReLU, and max pooling. An FFT is applied after each block to return to the frequency domain before the next convolution. The final block omits the FFT to retain the spatial representation, which is passed to an MLP for the same downstream uses as the FFT-Block.
}
\label{fig:blocks}
\end{figure}

Figure~\ref{fig:blocks} shows the Fast Fourier Transform Block (FFT-Block), an architecture designed to process frequency data directly. The first stage learns a frequency representation of the WSI and is implemented as an eight-layer CNN with $3 \times 3$ Conv2D, ReLU activation, and $2 \times 2$ MaxPool operations, without batch normalization. Batch normalization is excluded because it can introduce artifacts and compress feature values when applied to frequency data~\cite{chu2023rethinking}. Standard activation functions such as ReLU can also cause issues when applied to frequency data due to the zeroing of large negative values~\cite{chu2023rethinking}. However, our method addresses this by preprocessing frequency images into magnitude and phase representations in Equations~\ref{eq:magnitude} and \ref{eq:phase} which restricts the magnitude to positive values.

The frequency feature produced by the first-stage CNN contain large values and variance that are incompatible with fusion in conventional MIL-based architectures. To resolve this, we apply Min-Max normalization, which has been shown to provide a stable and effective approximation of neural network outputs without requiring standard deviation calculations~\cite{aboulatta2019stabilizing}. We find that Min-Max scaling not only enables frequency–spatial fusion but also improves overall performance as shown in Figure~\ref{fig:normalization}, which we attribute to more consistent feature distributions across examples, facilitating effective learning in subsequent stages.

The scaled feature is then fed to a second-stage MLP module whose output supports either standalone classification or fusion with MIL-based architectures. In the fusion setting, the MLP module projects the scaled feature into the MIL spatial feature space to integrate global context. Fusion is performed through element-wise addition, as illustrated in Figure~\ref{fig:diagram}. The frequency feature is added to each of CLAM’s~\cite{lu2021data} $N$ spatial features, enriching all patch-level representations with global context while preserving their relative differences. As a result, the attention scores remain unchanged, allowing MIL to preserve its patch-level weighting while incorporating the global context provided by the FFT-Block. A comparison of other fusion techniques is provided in Section~\ref{subsec:fusion}. The FFT-MIL framework can be summarized as follows.

\begin{equation}
\label{eq:FFT-Block}
O = \text{MLP}(\text{MinMax}(\text{CNN}(F_{\text{wsi}})))
\end{equation}
Here, \( F_{\text{wsi}} \) represents the frequency crop of a WSI. The first stage \text{CNN} module extracts features from \( F_{\text{wsi}} \), which are then scaled by a \text{MinMax} operation. Then, a second stage \text{MLP} (Multilayer Perceptron) produces \( O \), which can act as a global frequency feature for spatial fusion, or directly as a WSI label when performing standalone classification.

\begin{equation}
\hat{y} = \text{MIL}(O)
\end{equation}
The output \( O \) is then utilized by any MIL architecture to produce a WSI classification label \( \hat{y} \). Specifically, \( O \) is fused with a latent feature in MIL through addition, and the specific point of addition varies depending on the MIL architecture being used, as detailed in Section~\ref{subsec:comparison}.

\subsection{Frequency vs Patch-Based Processing.}
\label{subsec:complexity}
Our method operates in the frequency domain, where spatial frequencies are radially ordered by scale: low frequencies near the center capture coarse global structure, while high frequencies toward the edges represent fine detail. In natural images, including WSIs of resolution \( H \times W \), signal energy is heavily concentrated in the low-frequency region~\cite{song2023fourier}. The cumulative energy increases logarithmically with radial distance \( r \) from the spectrum center, following \( E(r) \propto \log(r) \)~\cite{ruderman1993statistics}. This property allows a small subset of low-frequency components to retain most of the image information. For example, retaining $50\%$ of total energy requires a radius \( r_{0.5} \propto (HW)^{1/4} \), corresponding to an input area \( A_{0.5} \propto (HW)^{1/2} \).

In contrast, current patch-based pipelines divide a WSI into non-overlapping patches of size \( P \times P \), yielding \( \frac{HW}{P^2} \) patches. Each patch is independently embedded into a \( D \)-dimensional feature vector using a pretrained encoder such as ResNet50~\cite{he2016deep}, where \( D \ll P^2 \). This results in a total input size of \( \mathcal{O}\left(\frac{HW}{P^2} \cdot D\right) \). Although this reduces the raw image size, individual features are spatially localized and do not capture global context. Moreover, MIL methods often face memory limitations when processing the full set of patch embeddings.

\begin{figure}[H]
    \centering
    \includegraphics[width=0.48\textwidth]{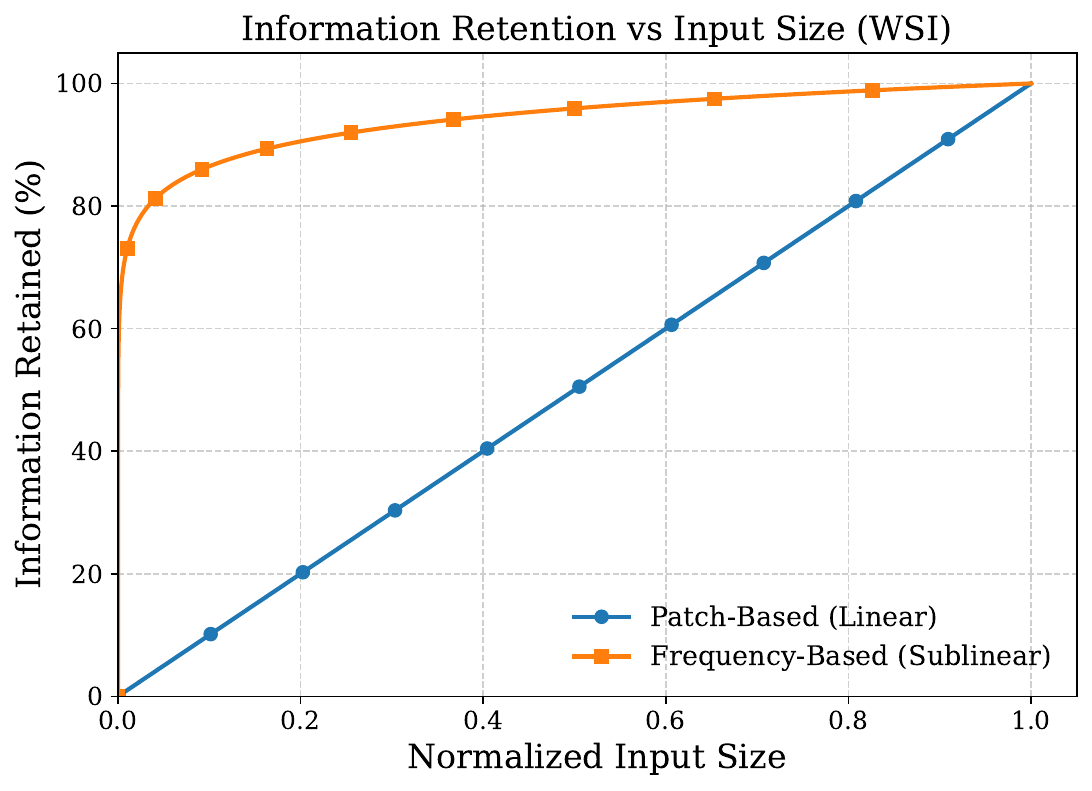}
    \caption{Information retention versus normalized input size for patch-based and frequency-based representations. A normalized input size of $1.0$ corresponds to full-image coverage. Patch-based input reflects the number of extracted patches multiplied by channel count and embedding dimensionality. Frequency-based input reflects the area of a radial crop in the Fourier domain. As shown, frequency-based inputs retain substantially more information at lower input sizes, highlighting their data efficiency in capturing global context compared to patch-based inputs.}
    \label{fig:complex}
\end{figure}

To compare frequency and patch-based inputs, we examine how much data is required to retain $50\%$ of the total WSI information. In patch-based methods, this corresponds to extracting and embedding half of all patches, which yields an input size of \( \mathcal{O}\left(\frac{HW}{2P^2} \cdot D\right) \). In contrast, the frequency-based approach achieves equivalent coverage with a radial area crop \( \mathcal{O}((HW)^{1/2}) \), without patch extraction or feature embedding. Figure~\ref{fig:complex} illustrates how input size scales with retained information. Patch-based representations grow linearly with resolution and provide only localized features. Frequency-based representations, on the contrary, offer global representations whose detail increases with crop size modeling of coarse structure in large resolution WSIs with less data. While they do not replace fine-grained patch-level detail, frequency-domain features provide a complementary global signal that addresses the context limitations of conventional MIL.

\section{Experiments}
\subsection{Dataset.}
\label{subsec:dataset}
FFT-MIL is evaluated on the WSI classification task across three different datasets: BRACS~\cite{brancati2022bracs} (536 images, 7 classes), IMP~\cite{neto2024interpretable} (826 images, 3 classes), and LUAD~\cite{gillette2020proteogenomic} (1,107 images, 2 classes). All slides are analyzed at $40\times$ magnification. Spatial streams use features from $256 \times 256$ patches extracted using CLAM's~\cite{lu2021data} preprocessing pipeline~\cite{lu2021data}, which removes whitespace and embeds tissue patches using a ResNet50~\cite{he2016deep} pretrained on ImageNet~\cite{deng2009imagenet}.

\subsection{Implementation Details.}
\label{subsec:Implementation Details.}
FFT-MIL is evaluated using three codebases and six unique MIL-based architectures, including CLAM's~\cite{lu2021data} implementation of the CLAM and MIL methods, ACMIL's~\cite{zhang2024attention} implementation of the ACMIL, ABMIL, and IBMIL methods, and DGR-MIL's~\cite{zhu2024dgr} implementation of the ABMIL and ILRA methods. We follow their implementation details and divide our datasets into $80\%-20\%$ train-test splits. Evaluation is standardized across all codebases to include accuracy, precision, recall, macro-averaged harmonic mean of precision and recall (F1 score), and macro-average one-vs-rest area under the curve (AUC) for each method. 

Model checkpoints are selected based on the macro-averaged F1 score. Compared to AUC-based selection, this yields an average improvement of $+4.5\%$ in F1 score and a $-1.3\%$ reduction in AUC, representing a favorable trade-off for class-balanced performance. The macro F1 score computes an unweighted average across all classes, mitigating the effects of class imbalance and reducing inter-method variance. To evaluate robustness in deployment-oriented settings, where majority-class performance has a greater influence on overall metrics, we repeat the experiments in Table~\ref{tab:results2} using weighted-averaged F1 score for model selection, observing an average gain of $+2.75\%$ in overall prediction accuracy.

The selected architectures encompass foundational and state-of-the-art MIL-based approaches for WSI classification. MIL~\cite{ilse2018attention} serves as the foundational framework, while CLAM~\cite{lu2021data} introduces a state-of-the-art improvement by combining a CNN-based feature extractor with an attention-based aggregator and instance-level clustering. The remaining methods, including ABMIL~\cite{ilse2018attention}, ACMIL~\cite{zhang2024attention}, IBMIL~\cite{lin2023interventional}, and ILRA~\cite{xiang2023exploring}, are also state-of-the-art, with several drawing conceptual inspiration from CLAM~\cite{lu2021data}. ABMIL~\cite{ilse2018attention} introduces a learnable attention pooling mechanism for instance weighting. ACMIL~\cite{zhang2024attention} enhances attention-based MIL through multi-branch attention and stochastic Top-K instance masking to promote diversity and prevent overfitting. IBMIL~\cite{lin2023interventional} incorporates interventional training and a learnable deconfounding module for causal adjustment. ILRA~\cite{xiang2023exploring} imposes low-rank constraints through specialized embedding and pooling modules to enable global instance interaction and improve generalization.

\subsection{Comparison with State-of-the-Art Methods.} 
\label{subsec:comparison}
To incorporate FFT-MIL with MIL-based methods, the FFT-Block's frequency feature is added with spatial features at a key point depending on the MIL-based method. The simplest case is the traditional MIL method~\cite{ilse2018attention} that processes the incoming patch features before performing an aggregation and classification. Here, the global frequency feature is aggregated after MIL processes the incoming patches. This introduces global context across all of the latent patch features, which can be utilized by the rest of the pipeline. The same key point is empirically determined for CLAM~\cite{lu2021data}, ABMIL~\cite{ilse2018attention}, IBMIL~\cite{lin2023interventional}, and ILRA~\cite{xiang2023exploring}, which consist of linear, attention, or attention pooling mechanisms for processing after given patch features. ACMIL~\cite{zhang2024attention} is the only MIL-based approach where we find that fusing the global frequency feature is most effective towards the end of the architecture and where we instead perform fusion before its classifier layer. We attribute this to ACMIL's~\cite{zhang2024attention} Stochastic Top-K Instance Masking module, which prevents overfitting by redistributing attention across multiple instances instead of focusing on a few dominant ones~\cite{zhang2024attention}.

The experimental results are presented in Table~\ref{tab:results}. We observe that FFT-MIL is most effective when combined with CLAM's approach~\cite{lu2021data}. We attribute this to adopting CLAM's~\cite{lu2021data} patch feature extraction process that is optimized for the method. Furthermore, we note that ILRA~\cite{xiang2023exploring} benefits the least from the global frequency feature. We attribute this to ILRA's low-rank attention pooling module that captures interactions among instances. Even so, the method still sees improvement from FFT-MIL due to being derived from the frequency domain, which utilizes the full WSI rather than a subset of patches. FFT-MIL improves the average performance of the adopted MIL-based methods by $+3.51\%$ in F1 score and $+1.51\%$ in AUC, demonstrating effective integration of frequency-derived global features with spatial models for enhanced WSI classification.

\begin{table}[H]
\centering
\caption{Evaluation of all methods as implemented by CLAM~\cite{lu2021data}, ACMIL~\cite{zhang2024attention}, and DGR-MIL~\cite{zhu2024dgr} on BRACS~\cite{brancati2022bracs}, LUAD~\cite{gillette2020proteogenomic}, and IMP~\cite{neto2024interpretable}, with Accuracy (ACC), Precision (PRE), Recall (REC), F1 score (F1), and Area Under the Curve (AUC). $\Delta$AUC and $\Delta$F1 denote the average relative percentage change achieved by integrating FFT-MIL into each baseline MIL method, including CLAM~\cite{lu2021data}, MIL~\cite{ilse2018attention}, ABMIL~\cite{ilse2018attention}, ACMIL~\cite{zhang2024attention}, IBMIL~\cite{lin2023interventional}, and ILRA~\cite{xiang2023exploring}, over the three datasets, BRACS~\cite{brancati2022bracs}, LUAD~\cite{gillette2020proteogenomic}, and IMP~\cite{neto2024interpretable}. Best results are marked in bold. Methods marked with “(Ours)” denote the integration of the proposed FFT-MIL framework into the corresponding baseline.}
\vspace{10pt}
\label{tab:results}
\scriptsize
\renewcommand{\arraystretch}{1.2}
\setlength{\tabcolsep}{3pt}
\begin{tabular}{@{}l
|>{\centering\arraybackslash}p{0.55cm}>{\centering\arraybackslash}p{0.55cm}>{\centering\arraybackslash}p{0.55cm}>{\centering\arraybackslash}p{0.55cm}>{\centering\arraybackslash}p{0.55cm}|
>{\centering\arraybackslash}p{0.55cm}>{\centering\arraybackslash}p{0.55cm}>{\centering\arraybackslash}p{0.55cm}>{\centering\arraybackslash}p{0.55cm}>{\centering\arraybackslash}p{0.55cm}|
>{\centering\arraybackslash}p{0.55cm}>{\centering\arraybackslash}p{0.55cm}>{\centering\arraybackslash}p{0.55cm}>{\centering\arraybackslash}p{0.55cm}>{\centering\arraybackslash}p{0.55cm}|
>{\centering\arraybackslash}p{0.75cm}
>{\centering\arraybackslash}p{0.75cm}@{}}
\toprule
\multirow{2}{*}{\textbf{Method}} &
\multicolumn{5}{c|}{ \textbf{BRACS~\cite{brancati2022bracs}}} &
\multicolumn{5}{c|}{ \textbf{IMP~\cite{neto2024interpretable}}} &
\multicolumn{5}{c|}{ \textbf{LUAD~\cite{gillette2020proteogenomic}}} &
\multirow{2}{*}{\textbf{ $\Delta$AUC}} &
\multirow{2}{*}{\textbf{ $\Delta$F1}} \\
&  \textbf{ACC} &  \textbf{PRE} &  \textbf{REC} &  \textbf{F1} &  \textbf{AUC}
&  \textbf{ACC} &  \textbf{PRE} &  \textbf{REC} &  \textbf{F1} &  \textbf{AUC}
&  \textbf{ACC} &  \textbf{PRE} &  \textbf{REC} &  \textbf{F1} &  \textbf{AUC}
& & \\
\midrule
\multicolumn{18}{c}{\textbf{CLAM~\cite{lu2021data}}} \\
\midrule
CLAM & 58.5 & \textbf{0.60} & 0.47 & 0.49 & 0.77 & 92.8 & 0.91 & 0.94 & 0.93 & 0.96 & 96.4 & 0.96 & 0.96 & 0.96 & 0.97 & -- & -- \\
CLAM (Ours) & \textbf{64.2} & \textbf{0.60} & \textbf{0.52} & \textbf{0.53} & \textbf{0.79} & \textbf{95.2} & 0.93 & \textbf{0.96} & \textbf{0.94} & 0.97 & \textbf{97.3} & 0.96 & \textbf{0.98} & \textbf{0.97} & 0.98 & \textcolor{darkgreen}{+1.5\%} & \textcolor{darkgreen}{+3.9\%} \\
MIL & 49.1 & 0.42 & 0.39 & 0.39 & 0.70 & 85.5 & 0.82 & 0.84 & 0.83 & 0.93 & 93.7 & 0.94 & 0.92 & 0.93 & 0.97 & -- & -- \\
MIL (Ours) & 52.8 & 0.49 & 0.42 & 0.42 & 0.72 & 91.6 & 0.90 & 0.91 & 0.90 & 0.95 & 94.6 & 0.93 & 0.95 & 0.94 & 0.97 & \textcolor{darkgreen}{+1.9\%} & \textcolor{darkgreen}{+5.9\%} \\
\midrule
\multicolumn{18}{c}{\textbf{ACMIL~\cite{zhang2024attention}}} \\
\midrule
ABMIL & 44.2 & 0.14 & 0.25 & 0.17 & 0.76 & 85.5 & 0.82 & 0.82 & 0.82 & 0.94 & 93.7 & 0.93 & 0.94 & 0.93 & 0.98 & -- & -- \\
ABMIL (Ours) & 46.7 & 0.14 & 0.26 & 0.18 & 0.78 & 85.5 & 0.82 & 0.84 & 0.82 & 0.94 & 93.7 & 0.92 & 0.95 & 0.93 & 0.98 & \textcolor{darkgreen}{+0.6\%} & \textcolor{darkgreen}{+1.8\%} \\
ACMIL & 42.3 & 0.15 & 0.24 & 0.17 & 0.67 & 78.3 & 0.84 & 0.65 & 0.64 & 0.91 & 94.6 & 0.94 & 0.94 & 0.94 & \textbf{0.99} & -- & -- \\
ACMIL (Ours) & 45.7 & 0.13 & 0.26 & 0.17 & 0.72 & 85.5 & 0.89 & 0.79 & 0.81 & 0.93 & 95.5 & 0.95 & 0.96 & 0.95 & \textbf{0.99} & \textcolor{darkgreen}{+\textbf{3.4\%}} & \textcolor{darkgreen}{\textbf{+9.7\%}} \\
IBMIL & 44.2 & 0.14 & 0.25 & 0.17 & 0.76 & 85.5 & 0.82 & 0.82 & 0.82 & 0.94 & 93.7 & 0.93 & 0.94 & 0.93 & 0.98 & -- & -- \\
IBMIL (Ours) & 46.7 & 0.14 & 0.26 & 0.18 & 0.78 & 85.5 & 0.82 & 0.84 & 0.82 & 0.94 & 93.7 & 0.92 & 0.95 & 0.93 & 0.98 & \textcolor{darkgreen}{+0.6\%} & \textcolor{darkgreen}{+1.8\%} \\
\midrule
\multicolumn{18}{c}{\textbf{DGR-MIL~\cite{zhu2024dgr}}} \\
\midrule
ABMIL & 60.4 & \textbf{0.60} & 0.45 & 0.45 & 0.74 & 91.6 & 0.90 & 0.92 & 0.91 & 0.97 & 95.5 & 0.95 & 0.95 & 0.95 & 0.97 & -- & -- \\
ABMIL (Ours) & 60.4 & 0.56 & 0.47 & 0.47 & 0.75 & 92.8 & 0.91 & 0.94 & 0.92 & 0.97 & 96.4 & 0.96 & 0.97 & 0.96 & \textbf{0.99} & \textcolor{darkgreen}{+1.2\%} & \textcolor{darkgreen}{+2.2\%} \\
ILRA & 52.8 & 0.52 & 0.47 & 0.47 & 0.74 & 92.8 & 0.92 & 0.90 & 0.91 & \textbf{0.98} & 96.4 & 0.96 & 0.96 & 0.96 & \textbf{0.99} & -- & -- \\
ILRA (Ours) & 54.7 & 0.45 & 0.46 & 0.45 & 0.77 & 94.0 & \textbf{0.94} & 0.92 & 0.93 & \textbf{0.98} & \textbf{97.3} & \textbf{0.97} & 0.97 & \textbf{0.97} & \textbf{0.99} & \textcolor{darkgreen}{+1.4\%} & \textcolor{darkred}{-0.7\%} \\
\bottomrule
\end{tabular}
\end{table}

\begin{figure}[H]
\includegraphics[width=\textwidth]{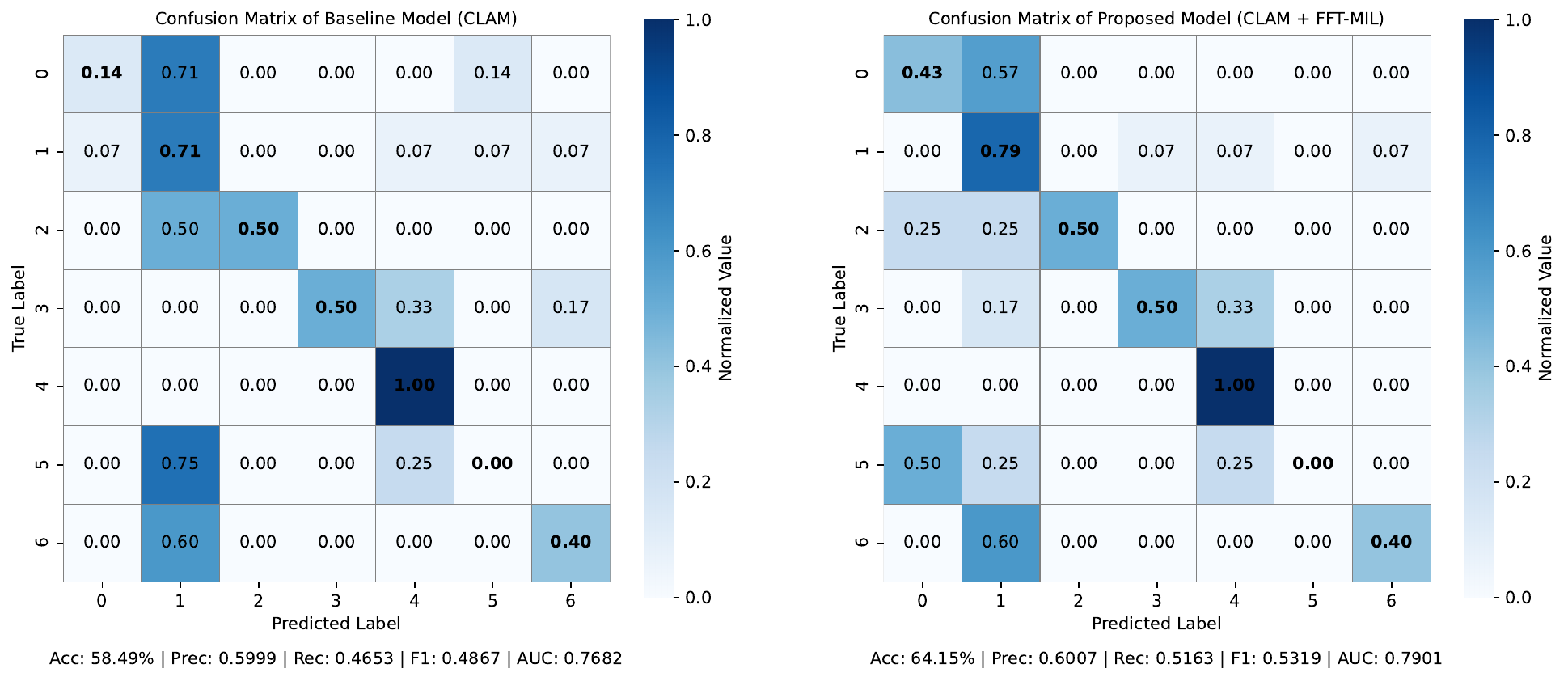}
\caption{Normalized confusion matrices comparing the classification performance of the baseline CLAM model (left) and the proposed FFT-MIL model (right) on BRACS~\cite{brancati2022bracs}. Each matrix illustrates the normalized distribution of true versus predicted class labels. Summary metrics below each matrix include Accuracy (Acc), Precision (Prec), Recall (Rec), F1 score (F1), and Area Under the Curve (AUC). FFT-MIL demonstrates improved predictive performance as indicated by higher diagonal values in the confusion matrix.}
\label{fig:matrices}
\end{figure}

In Figure~\ref{fig:matrices} we compare the normalized confusion matrices of the baseline CLAM~\cite{lu2021data} model and FFT-MIL on the BRACS~\cite{brancati2022bracs} dataset to assess class-specific performance. FFT-MIL shows improved prediction across multiple classes, including classes 0 and 1. In addition, class 5 shows a more balanced distribution of predictions, suggesting improved handling of underrepresented categories. These improvements are reflected in higher accuracy, precision, recall, F1 score, and AUC, indicating more consistent and robust classification performance.

\begin{figure}[H]
\includegraphics[width=\textwidth]{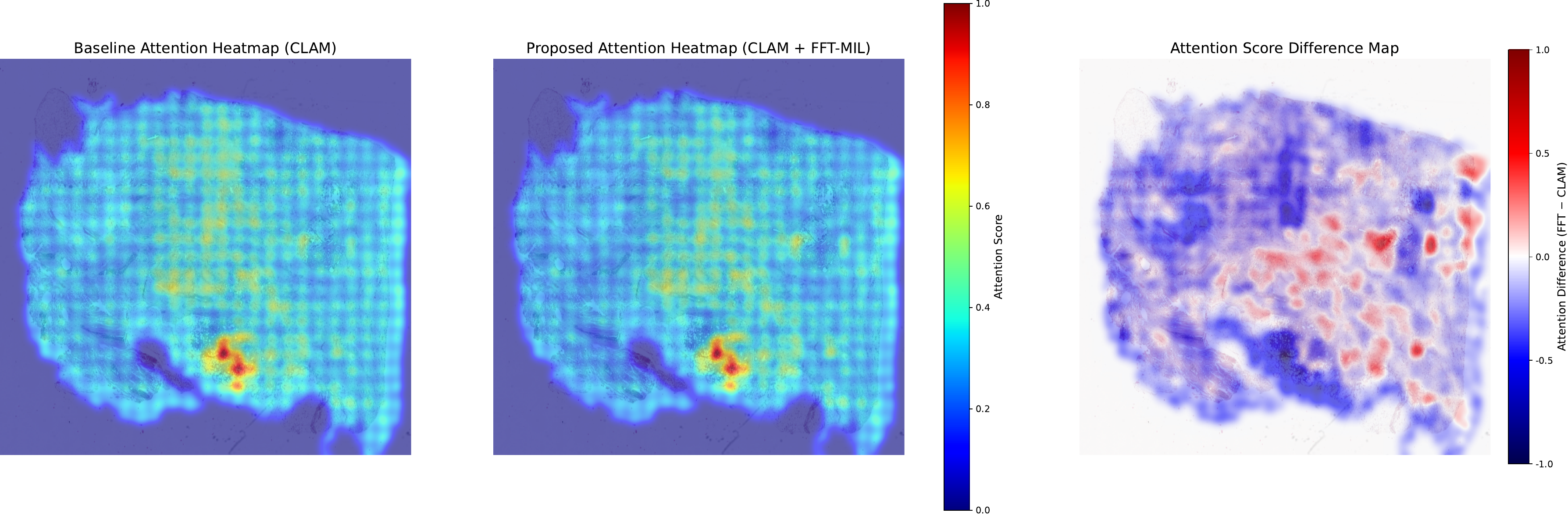}
\caption{Attention heatmaps for a representative WSI from the BRACS~\cite{brancati2022bracs} dataset. The baseline CLAM model's attention scores (left) are compared with those from the proposed FFT-MIL model (center). The rightmost panel shows the difference between the two attention scores, highlighting regions where the proposed model assigns higher (red) or lower (blue) attention relative to the baseline. The difference map illustrates that FFT-MIL yields more localized and concentrated attention compared to the baseline.}
\label{fig:heatmaps}
\end{figure}

In Figure~\ref{fig:heatmaps} we compare attention heatmaps from the baseline CLAM~\cite{lu2021data} and our proposed FFT-MIL model on a representative WSI from BRACS~\cite{brancati2022bracs} to investigate the spatial impact of frequency-domain integration. Because both models visually highlight similar regions, we include a third heatmap showing the pixel-wise difference to localize areas of divergence in attention. The baseline CLAM exhibits broadly dispersed attention, reflecting a lack of spatial precision and limited use of global context. In contrast, FFT-MIL produces more concentrated attention, supported by a $16.0\%$ reduction in entropy and a $23.2\%$ increase in standard deviation, indicating a sharper and more selective focus. Furthermore, a center-of-mass shift of $317.7$ pixels confirms a measurable spatial adjustment. These findings demonstrate that FFT-MIL maintains alignment with the primary semantic regions identified by CLAM~\cite{lu2021data}, while producing more spatially selective and concentrated attention distributions.

\begin{figure}[H]
\includegraphics[width=\textwidth]{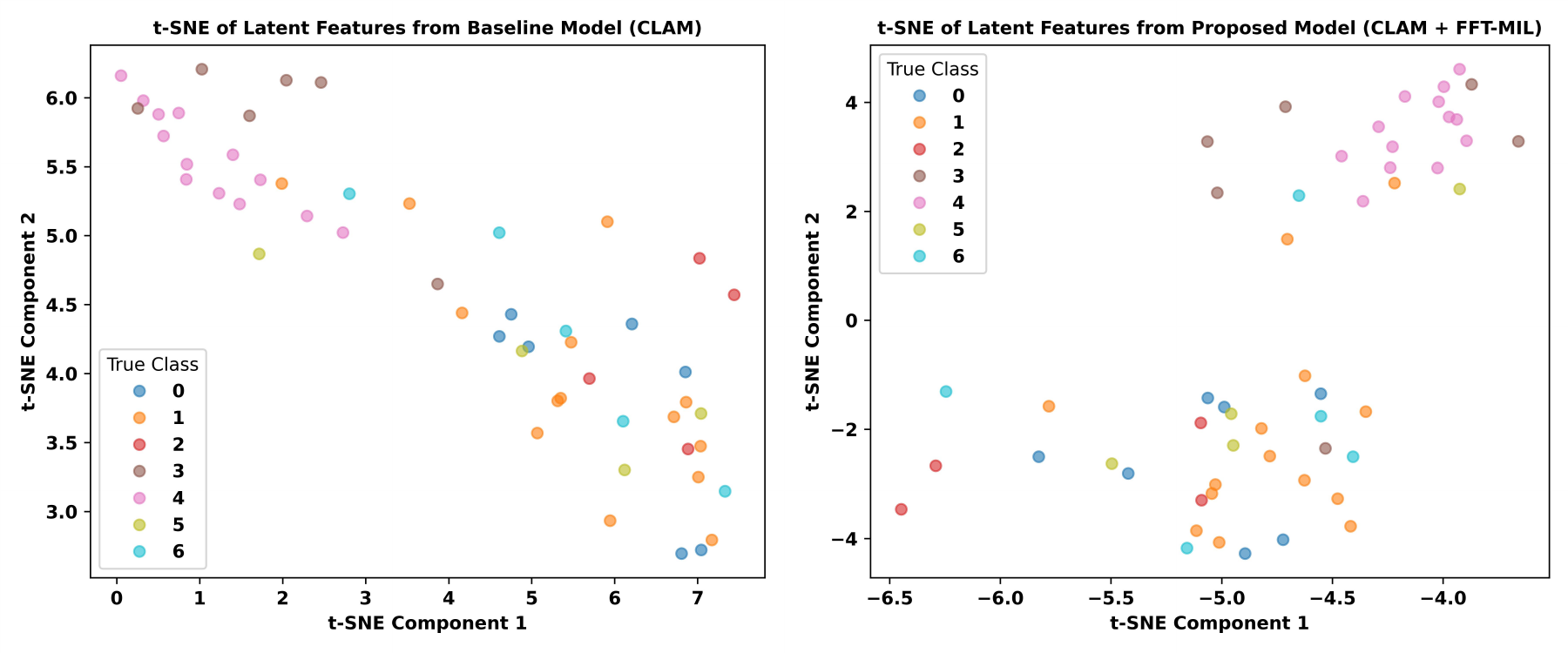}
\caption{t-SNE visualizations of latent features extracted from the baseline CLAM~\cite{lu2021data} model and the proposed FFT-MIL model on the BRACS~\cite{brancati2022bracs} dataset. Each point represents a WSI and is colored by its ground truth class label. FFT-MIL produces more compact and well-separated clusters in the embedded space, indicating improved feature discriminability enabled by frequency-domain integration.}
\label{fig:tsne}
\end{figure}

In Figure~\ref{fig:tsne}, we compare t-SNE visualizations of latent features from the CLAM baseline and our proposed FFT-MIL model on the BRACS~\cite{brancati2022bracs} dataset to assess representation quality. Visually, FFT-MIL exhibits tighter intra-class clustering and greater inter-class separation. Quantitatively, FFT-MIL improves 2D k-NN classification accuracy by $7.4\%$ and macro F1 score by $23.3\%$, confirming the increased discriminability and class consistency of the learned features. These results demonstrate that FFT-MIL enhances the structure and separability of the latent space, supporting more interpretable and class-aware representations.

\section{Ablation Study}
The following ablation studies evaluate the design choices and trade-offs of FFT-MIL. Section~\ref{subsec:abl1} investigates how the FFT-Block learns from frequency representations of WSIs, analyzing spectral components, informative regions, crop size, downsampling effects, and comparing different normalization techniques within the FFT-Block. Section~\ref{subsec:fftdesigns} compares our FFT-Block design with prior frequency architectures and examines how their design choices affect performance. Section~\ref{subsec:fusion} evaluates alternative strategies for fusing spatial and frequency features. Section~\ref{subsec:representations} tests other compressed transformation methods within our framework. Section~\ref{subsec:cost} reports computational efficiency relative to spatial and multiscale baselines. Section~\ref{subsec:abl2} contrasts frequency-only and spatial-only models to highlight their complementary roles. Finally, Section~\ref{subsec:abl3} analyzes the robustness of FFT-MIL to class imbalance.

\subsection{Analysis of Frequency Representations and Preprocessing}
\label{subsec:abl1}
\begin{figure}[H]
\includegraphics[width=\textwidth]{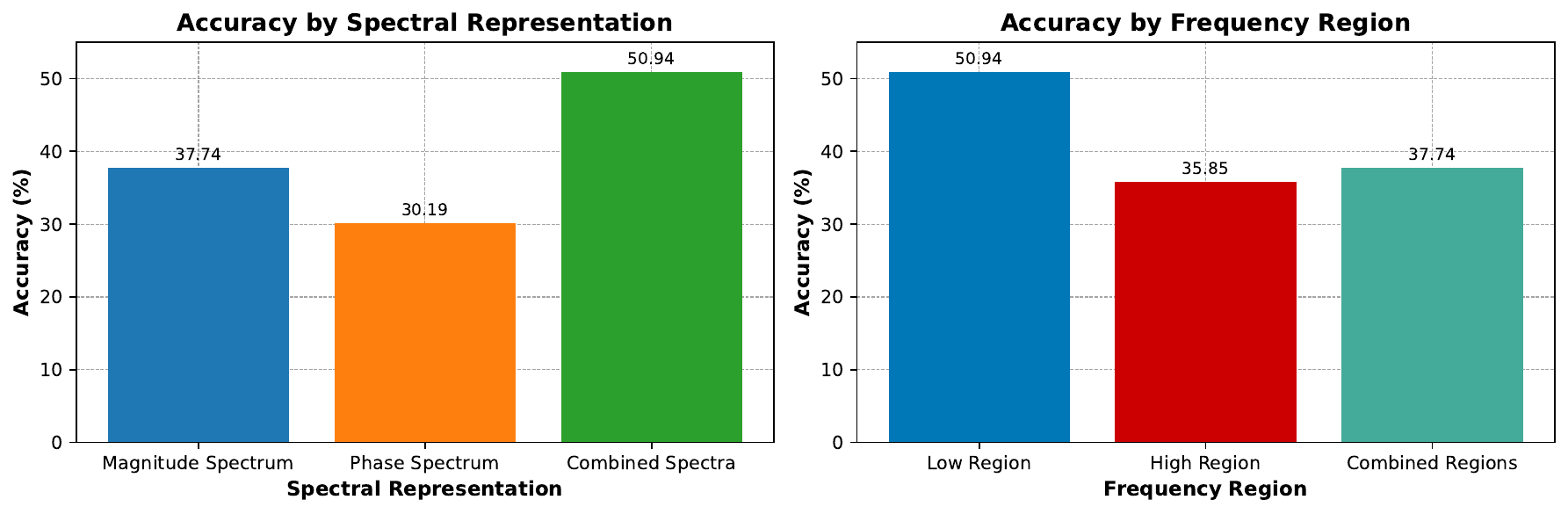}
\caption{Classification accuracy of the proposed FFT-Block on the BRACS~\cite{brancati2022bracs} dataset using different spectral inputs (left) and frequency regions (right). Experiments were conducted on $2048 \times 2048$ WSI frequency-domain crops. The magnitude spectrum is the most informative individual component, while combining magnitude and phase yields the highest performance by enabling a complete representation of the frequency image. Low-frequency regions contribute most to the effectiveness of the proposed FFT-Block, consistent with their higher energy concentration in the frequency domain.}
\label{fig:spectra_region}
\end{figure}

We begin by evaluating how best to leverage frequency representations of WSIs, focusing on both spectral components and spatial regions. First, we test magnitude and phase representations extracted from a low-frequency center crop of WSIs. The magnitude spectrum primarily encodes intensity information, while the phase spectrum captures structural details~\cite{huang2022deep}. As shown in Figure~\ref{fig:spectra_region}, the magnitude spectrum alone is more informative than the phase spectrum for WSI analysis. However, since both are required for a complete frequency representation, as described in Section~\ref{subsec:preprocessing}, their combination results in the best performance.

Spectrally, low frequencies correspond to slow intensity variations and capture global structure, whereas high frequencies encode rapid changes such as edges~\cite{shaikh2016analysis}. To identify the most informative regions, we analyze center crops taken before and after zero-frequency centering, as visualized in Figure~\ref{fig:preprocessing}. Before shifting, the crop corresponds to high-frequency content; after shifting, it captures low-frequency components. We also evaluate a combined setting where both are concatenated and jointly learned. As shown in Figure~\ref{fig:spectra_region}, low-frequency regions are more effective in capturing global context, consistent with their higher energy concentration and importance in image reconstruction~\cite{pandey2015image}. Notably, using only low frequencies outperforms the combined setting, suggesting that more advanced fusion strategies in our FFT-Block may be required to fully leverage high-frequency information.

\begin{figure}[H]
\includegraphics[width=\textwidth]{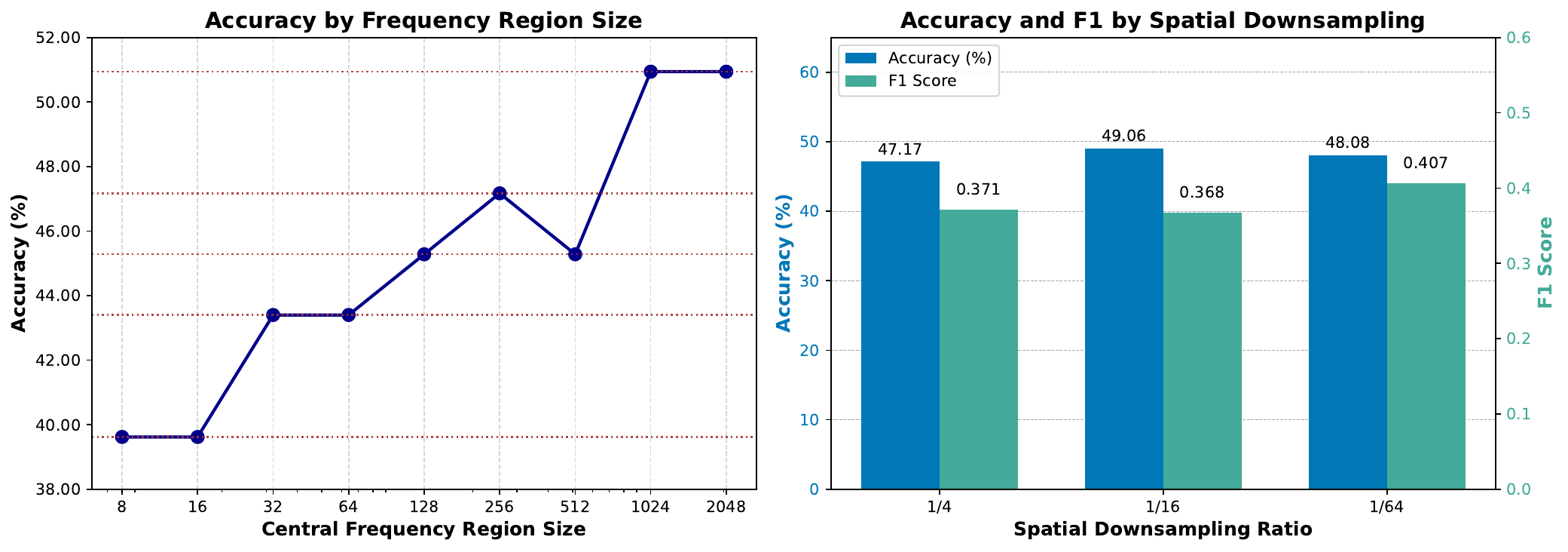}
\caption{Classification performance of the proposed FFT-Block on the BRACS~\cite{brancati2022bracs} dataset under varying (left) frequency crop sizes and (right) image resolutions, based on WSI frequency representations. Performance improves with larger frequency crops, reflecting the increased information content captured. In contrast, WSI downsampling does not degrade performance, because the corresponding frequency crop encompasses a greater portion of the original image.}
\label{fig:down_sizes}
\end{figure}

Next, we evaluate the impact of the frequency crop size and initial WSI downsampling on performance. As shown in Figure~\ref{fig:down_sizes}, increasing the size of the low-frequency center crop leads to better performance, consistent with previous findings that larger low-frequency regions retain more image information and improve reconstruction quality~\cite{ruderman1993statistics}. As the center crop expands, it progressively covers more mid-frequency components which, although less energy-dense than the central low-frequency components, provide complementary information that enhances performance. 

Surprisingly, increasing the downsampling factor of the WSI prior to preprocessing, as shown in Figure~\ref{fig:down_sizes}, does not reduce performance, despite the expected degradation in visual detail~\cite{trentacoste2011blur}. This suggests that downsampling may enhance the representational efficiency of a fixed-size center crop by allowing it to capture a larger portion of the original image. We hypothesize a tradeoff between crop size and spatial downsampling that may be jointly optimized. This tradeoff is particularly important in practice, as crop size scales quadratically with memory requirements during training.

\begin{figure}[H]
    \centering
    \includegraphics[width=0.48\textwidth]{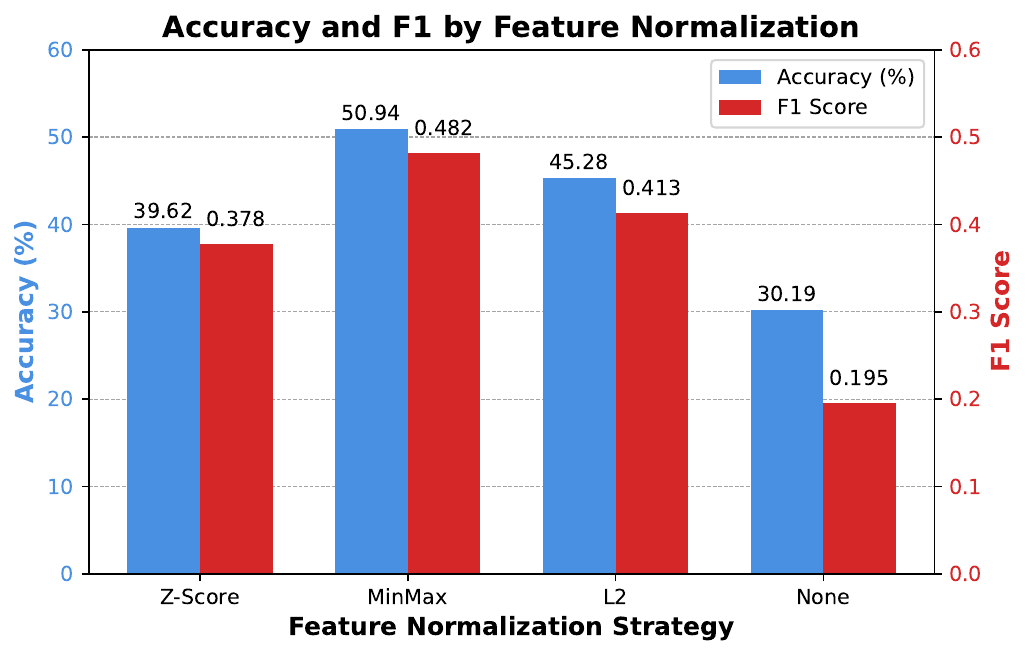}
    \caption{Accuracy and F1 scores of our proposed FFT-Block on BRACS~\cite{brancati2022bracs} using WSI frequency representations under different feature normalization methods: Z-Score, Min-Max, L2, and None (no normalization). All normalization methods improve performance by standardizing the distribution of frequency features, which facilitates more stable and effective learning. Min-Max normalization yields the highest gains by preserving relative feature structure while constraining values to a fixed range.}
    \label{fig:normalization}
\end{figure}

Finally, we evaluate which normalization technique is the most effective for the FFT-Block in Figure~\ref{fig:normalization}. Specifically, as described in Equation~\ref{eq:FFT-Block}, we apply normalization to the output of the CNN module before being fed to a MLP to allow for spatial-frequency feature fusion. We compare the L2, Z-Score, and Min-Max techniques because they are widely used and conceptually distinct~\cite{cabello2023impact}. L2 normalization scales entire features to unit length~\cite{wang2017normface}, Z-Score centers around mean 0 with unit variance scaling, and Min-Max scales features to a fixed range~\cite{patro2015normalization}. We find Min-Max normalization outperforms other techniques, and believe it is due to preservation of the relative structure of frequency features while constraining their range to $[0, 1]$. In contrast, Z-Score normalization introduces instability due to the heavy-tailed distribution of FFT features, and L2 normalization removes meaningful activation strength by flattening differences in overall frequency intensity.

\subsection{Analysis of Frequency Architecture Design Choices}
\label{subsec:fftdesigns}
To assess the impact of design decisions from prior frequency architectures, we evaluate their effect on our proposed FFT-Block and FFT-Block Vanilla, whose architectures are visualized in Figure~\ref{fig:blocks}. The results of this evaluation are summarized in Table~\ref{tab:ablation_fft_block}, 
where individual designs are denoted by letters (\textbf{A}, \textbf{B}, …, \textbf{H}) 
for ease of reference in the discussion.

FFT-Block Vanilla's design (\textbf{A}) follows prior works~\cite{song2023fourier, pathak2022fourcastnet}, which apply a single learnable layer before performing all subsequent operations in the spatial domain. In our experiments, we replace FFT-Block Vanilla’s layers and operations with complex versions, as leveraged in certain prior works~\cite{wang2024freqformer, xu2024dual}, which normalize the real and imaginary parts independently and allow frequency inputs to be processed directly. Extending FFT-Block Vanilla, most frequency architectures instead apply ReLU activation directly in the frequency domain (\textbf{B}) before converting features to the spatial domain, leading to significantly higher performance~\cite{li2020fourier, rippel2015spectral, zheng2024fouriermil, wang2024freqformer, xu2024dual, chu2023rethinking, zhang2022swinfir, huang2022deep}. Further, some architectures~\cite{xu2024dual, zhang2022swinfir, huang2022deep} replace ReLU with Leaky ReLU (\textbf{C}) to better handle negative frequency values, leading to additional performance gains. Finally, some architectures~\cite{chi2020fast, paing2023adenoma, suvorov2022resolution} include both batch normalization and activation in the frequency domain (\textbf{D}), offering a modest improvement over using activation alone.

Next, we compare the performance of our proposed FFT-Block (\textbf{E}), as described in Section~\ref{subsec:block}, with prior frequency architectures. Compared to FFT-Block Vanilla, our method achieves substantially higher performance, particularly in F1 score. We then evaluate a variant of the FFT-Block that omits separating frequency inputs into magnitude and phase, and instead processes complex-valued inputs directly with complex convolutional layers (\textbf{F}). This does not significantly affect performance. Building on the complex-valued variant, we replace Min-Max normalization with an iFFT (\textbf{G}) to convert the frequency feature to the spatial domain, as is standard in frequency architectures~\cite{li2020fourier, suvorov2022resolution, nair2020fast, chu2023rethinking, huang2022deep, zhang2022swinfir, wen2022u, pathak2022fourcastnet, paing2023adenoma, zheng2024fouriermil}. Differing from prior work~\cite{xu2024dual}, which applied at most three layers directly to a frequency input, this design employs an eight-layer CNN before transforming the representation to the spatial domain with an iFFT, enabling a deeper and more expressive representation. However, normalization is absent in this variant, as the Min-Max operation is replaced by an iFFT that directly projects the CNN output into the spatial domain. To address this, batch normalization is incorporated into the CNN layers (H), but performance degrades because normalization is applied to early frequency features before a mature encoding is established. Finally, to more closely mirror our FFT-Block which applies batch normalization only at the bottleneck after the CNN has produced a complete feature representation, we add a single batch normalization operation after the iFFT (I). The resulting deep iFFT approach significantly outperforms FFT-Block Vanilla while incurring only a minor performance drop relative to our proposed FFT-Block.

Overall, our ablation study shows that frequency architectures benefit from applying activation directly in the frequency domain, as demonstrated in prior works. Increasing network depth, which has not been explored in prior frequency architectures, produces stronger frequency representations. Although batch normalization is standard in spatial-domain encoders, applying it to frequency data degrades performance due to the high variance inherent in frequency-domain representations. Our proposed FFT-Block, which employs Min-Max normalization in place of the iFFT used in prior works, achieves the best overall results.

\begin{table}[H]
\centering
\caption{Architectural designs are evaluated by replacing the FFT-Block in FFT-MIL (Figure~\ref{fig:diagram}) on the BRACS~\cite{brancati2022bracs} dataset. Each design is labeled by a reference letter (\textbf{A–I}). Metrics reported are weighted-averaged F1 score (F1) and Area Under the Curve (AUC), with $\Delta$ values denoting relative change compared to the respective FFT-Block Vanilla or FFT-Block baseline. ReLU and Leaky ReLU indicate that activation functions moved to the frequency domain. Batch Norm denotes normalization, where \checkmark$^{L}$ integrates it into CNN layers and \checkmark$^{B}$ applies a single normalization in the spatial domain. Complex Layers indicates the use of complex-valued convolutions, while iFFT denotes replacing the Min-Max normalization of the FFT-Block with an inverse FFT.}
\vspace{10pt}
\label{tab:ablation_fft_block}
\scriptsize
\renewcommand{\arraystretch}{1.5}
\setlength{\tabcolsep}{6pt}
\begin{tabular}{@{}c l c c c c c c c@{}}
\toprule
\textbf{Design} & \textbf{Architecture} & \textbf{ReLU} & \textbf{Leaky ReLU} & \textbf{Batch Norm} & \textbf{F1} & \textbf{AUC} & \textbf{$\Delta$F1} & \textbf{$\Delta$AUC} \\
\midrule
A & FFT-Block Vanilla &            &            &            & 0.227 & 0.576 & -- & -- \\
B & FFT-Block Vanilla & \checkmark &            &            & 0.329 & \textbf{0.733} & \textcolor{darkgreen}{+44.91\%} & \textcolor{darkgreen}{+27.19\%} \\
C & FFT-Block Vanilla &            & \checkmark &            & \textbf{0.367} & 0.725 & \textcolor{darkgreen}{+61.66\%} & \textcolor{darkgreen}{+25.82\%} \\
D & FFT-Block Vanilla & \checkmark &            & \checkmark$^{L}$ & 0.335 & 0.791 & \textcolor{darkgreen}{+47.73\%} & \textcolor{darkgreen}{+37.21\%} \\
\midrule
\textbf{Design} & \textbf{Architecture} & \textbf{Complex Layers} & \textbf{iFFT} & \textbf{Batch Norm} & \textbf{F1} & \textbf{AUC} & \textbf{$\Delta$F1} & \textbf{$\Delta$AUC} \\
\midrule
E & FFT-Block (Ours) &            &            &            & \textbf{0.525} & 0.815 & -- & -- \\
F & FFT-Block & \checkmark &            &            & 0.521 & 0.817 & \textcolor{darkred}{-0.72\%} & \textcolor{darkgreen}{+0.23\%} \\
G & FFT-Block & \checkmark & \checkmark &            & 0.485 & \textbf{0.820} & \textcolor{darkred}{-7.71\%} & \textcolor{darkgreen}{+0.65\%} \\
H & FFT-Block & \checkmark & \checkmark & \checkmark$^{L}$ & 0.468 & 0.822 & \textcolor{darkred}{-10.84\%} & \textcolor{darkgreen}{+0.79\%} \\
I & FFT-Block & \checkmark & \checkmark & \checkmark$^{B}$ & 0.511 & 0.811 & \textcolor{darkred}{-2.70\%} & \textcolor{darkred}{-0.54\%} \\
\bottomrule
\end{tabular}
\end{table}

\subsection{Comparison of Fusion Strategies}
\label{subsec:fusion}
An important consideration is the effective fusion of spatial and frequency features, as implemented in our method shown in Figure~\ref{fig:diagram}. Table~\ref{tab:feature_fusion_comparison} compares several commonly used fusion strategies, including element-wise addition, element-wise multiplication, concatenation, and cross-attention~\cite{zhao2024deep}. Prior frequency architectures primarily employ addition~\cite{li2020fourier, chi2020fast, song2023fourier, pathak2022fourcastnet} and concatenation~\cite{wang2024freqformer, xu2024dual, paing2023adenoma, chu2023rethinking, zhang2022swinfir, suvorov2022resolution, huang2022deep}, with concatenation being the most widely adopted.

Our method uses addition to fuse a single frequency feature with each spatial feature generated from spatial patches. Doing so shifts every spatial feature equally and does not affect the attention scores generated by CLAM's~\cite{lu2021data} attention backbone. We next evaluate multiplication of the frequency feature with each patch feature, but find it ineffective. Most commonly, concatenation is employed in frequency architectures, which we implement by combining a copy of the frequency feature with each spatial feature and applying a linear projection layer to reduce each feature to its original size of $(N, 512)$ for further processing. This also results in significantly worse performance than addition. Finally, we apply cross-attention fusion, where the FFT-derived feature serves as the query and the patch features act as keys and values. This introduces a frequency-guided attention map, which is combined with CLAM’s~\cite{lu2021data} instance attention through a learnable softmax. The resulting fused attention is used to pool the patch features into a global representation, which is then added back to all patch features through a residual update, thereby injecting frequency context into the patch embeddings. 

Among the evaluated strategies, fusing frequency and spatial features through addition yields substantial improvements over multiplication and concatenation. Although concatenation is the most widely used strategy in prior frequency architectures, we show that our proposed addition fusion is effective and better suited to our method. Cross-attention provides a further gain beyond addition. Whereas addition integrates global context only into CLAM’s~\cite{lu2021data} spatial features, cross-attention directly modulates the attention scores, demonstrating the benefit of guiding patch weighting with global information. These findings suggest that specialized cross-attention designs hold strong potential for advancing frequency–spatial integration in MIL-based approaches.

\begin{table}[H]
\centering
\caption{Comparison of feature fusion strategies for integrating frequency and spatial features in FFT-MIL on the BRACS~\cite{brancati2022bracs} dataset. Metrics reported are weighted-averaged F1 score (F1) and Area Under the Curve (AUC). $\Delta$ values denote relative change compared to the baseline Element-Wise Addition. Fusion techniques include Element-Wise Multiplication, Concatenation, and Cross-Attention, where the FFT-derived global feature modulates patch-level spatial features through different integration mechanisms.}
\vspace{10pt}
\label{tab:feature_fusion_comparison}
\scriptsize
\renewcommand{\arraystretch}{1.5}
\setlength{\tabcolsep}{6pt}
\begin{tabular}{@{}l c c c c@{}}
\toprule
\textbf{Fusion Technique} & \textbf{F1} & \textbf{AUC} & \textbf{$\Delta$F1} & \textbf{$\Delta$AUC} \\
\midrule
Element-Wise Addition (Ours)        & 0.525 & 0.815 & -- & -- \\
Cross-Attention              & 0.563 & 0.822 & \textcolor{darkgreen}{+7.18\%} & \textcolor{darkgreen}{+0.79\%} \\
Element-Wise Multiplication  & 0.465 & 0.789 & \textcolor{darkred}{-11.52\%} & \textcolor{darkred}{-3.18\%} \\
Concatenation                & 0.451 & 0.793 & \textcolor{darkred}{-14.19\%} & \textcolor{darkred}{-2.70\%} \\
\bottomrule
\end{tabular}
\end{table}

\subsection{Evaluation of Alternative Compressed Representations}
\label{subsec:representations}
To evaluate the effectiveness of the proposed Fast Fourier Transform for extracting a compressed image representation, shown in Figure~\ref{fig:preprocessing}, it is compared with common compression methods~\cite{dhawan2011review} such as the Real Fast Fourier Transform, Discrete Cosine Transform, and Discrete Wavelet Transform. The results are presented in Table~\ref{tab:frequency_transform_comparison}.

The Real Fast Fourier Transform (rFFT) is adopted in many prior frequency architecture approaches~\cite{wang2024freqformer, cakaj2023spectral, chu2023rethinking, zhang2022swinfir, suvorov2022resolution} for its exploitation of the Hermitian symmetry of real-valued images, enabling compact frequency representations using only half of the spectrum. Its preprocessing is identical to our FFT in Section~\ref{subsec:preprocessing}, except that the low-frequency crop is applied to the top-left region, where the low-frequency components are concentrated. We observe a decrease in performance, suggesting that the negative frequency components preserved by the FFT may contribute valuable information. In addition, because the rFFT discards the negative spectrum, it also prevents energy centering around the spectrum origin, and these two factors together may underlie the reduced performance.

Next, we evaluate the Discrete Cosine Transform (DCT). Closely related to the FFT, the DCT employs real-valued cosine bases derived from an even-symmetric extension and is widely used in image compression, most notably JPEG~\cite{dhawan2011review, ahmed2006discrete}. Since low-frequency content is concentrated in the top-left of the coefficient map, the $(2048, 2048)$ crop is taken from this region. The DCT produces only real-valued amplitudes without an explicit phase component and exhibits the same issue as real FFT coefficients, where large positive and negative values lead ReLU to suppress negative responses~\cite{chu2023rethinking}. To mitigate this, we use the absolute values of the coefficients, which improves the F1 score from $0.343$ to $0.468$ at the cost of information loss. Despite this gain, the DCT still underperforms compared to our proposed FFT preprocessing, which we attribute to its reliance on cosine-only bases and the absence of phase information, making it less expressive than the full FFT representation. However, because the DCT yields three channels instead of the six required to represent magnitude and phase pairs in our proposed FFT representation, it would allow a larger crop under the same computational budget. Further exploration of the DCT remains a potential direction for future works.

Finally, we compare the Discrete Wavelet Transform (DWT) due to its widespread acceptance in signal processing~\cite{dhawan2011review}. The DWT decomposes an image into four spatial sub-bands (LL, LH, HL, HH), where the LL component captures coarse low-frequency structure which we use for comparison. Since the LL sub-band reduces an input image by only one quarter, we interpolate it to the expected size of $(2048, 2048)$. As this is a spatial rather than frequency representation, we adapt our FFT-Block by replacing it with a conventional CNN using batch normalization and removing the Min-Max operation, which is not standard in vision architectures. This representation yields substantially lower performance, which is expected given that the FFT-Block was not designed for spatial inputs. Nonetheless, we speculate that the DWT could be effective in conjunction with a multiscale architecture that leverages all four sub-bands for future works.

When evaluated against alternative compressed representations, our method performs best with the FFT, consistent with its design. These comparisons highlight the limitations of directly substituting other transforms and provide insights into how methods might be tailored to better exploit rFFT, DCT, or DWT representations. More broadly, this analysis underscores the importance of aligning architectural choices with the properties of the underlying transform when developing frequency-based methods.

\begin{table}[H]
\centering
\caption{Comparison of feature fusion strategies for integrating frequency and spatial features (visualized in Figure~\ref{fig:diagram}) in FFT-MIL on the BRACS~\cite{brancati2022bracs} dataset. Metrics reported are weighted-averaged F1 score (F1) and Area Under the Curve (AUC), with $\Delta$ values denoting relative change compared to the baseline Element-Wise Addition. Evaluated techniques include Element-Wise Multiplication, Concatenation, and Cross-Attention, where the FFT-derived global feature modulates patch-level spatial features through different integration mechanisms.}

\vspace{10pt}
\label{tab:frequency_transform_comparison}
\scriptsize
\renewcommand{\arraystretch}{1.5}
\setlength{\tabcolsep}{6pt}
\begin{tabular}{@{}l c c c c@{}}
\toprule
\textbf{Method} & \textbf{F1} & \textbf{AUC} & \textbf{$\Delta$F1} & \textbf{$\Delta$AUC} \\
\midrule
Fast Fourier Transform (Ours)       & 0.525 & 0.815 & -- & -- \\
Real Fast Fourier Transform  & 0.465 & 0.808 & \textcolor{darkred}{-11.37\%} & \textcolor{darkred}{-0.87\%} \\
Discrete Cosine Transform        & 0.468 & 0.820 & \textcolor{darkred}{-10.82\%} & \textcolor{darkgreen}{+0.59\%} \\
Discrete Wavelet Transform       & 0.288 & 0.616 & \textcolor{darkred}{-45.12\%} & \textcolor{darkred}{-24.45\%} \\
\bottomrule
\end{tabular}
\end{table}

\subsection{Computational Efficiency Analysis}
\label{subsec:cost}
To evaluate the computational cost of our method, we compare it with the CLAM~\cite{lu2021data} baseline in Table~\ref{tab:resource_comparison}. We observe a modest increase in memory usage, which is attributed to the relatively small size of the single frequency crop compared to the numerous patches required in spatial MIL methods. Training runtime is also longer, reflecting the additional processing introduced by the frequency branch. Model parameters increase substantially due to the layer sizes chosen for the FFT-Block, and Table~\ref{tab:fft_variants_comparison} further compares downstream performance when the FFT-Block is configured with reduced layer sizes. Overall, these computational costs are expected, as FFT-MIL introduces an additional frequency branch, shown in Figure~\ref{fig:diagram}, which leads to improved downstream performance, as reported in Table~\ref{tab:results}.

\begin{table}[H]
\centering
\caption{Resource comparison between the baseline CLAM~\cite{lu2021data} and FFT-MIL (Figure~\ref{fig:diagram}) on the BRACS~\cite{brancati2022bracs} dataset. Reported metrics include total runtime, CPU memory, GPU memory, inference throughput (samples/s), and model parameters. Percentage difference is computed relative to the baseline CLAM~\cite{lu2021data} implementation.}
\vspace{10pt}
\label{tab:resource_comparison}
\scriptsize
\renewcommand{\arraystretch}{1.5}
\setlength{\tabcolsep}{6pt}
\begin{tabular}{@{}l c c c@{}}
\toprule
\textbf{Metric} & \textbf{CLAM} & \textbf{FFT-MIL (Ours)} & \textbf{Percentage Difference} \\
\midrule
Runtime (hours)          & 15.54 & 21.20 & \textcolor{darkred}{+36.43\%} \\
CPU Memory (MB)       & 1169  & 1354  & \textcolor{darkred}{+15.82\%} \\
GPU Memory (MB)       & 2134  & 2673  & \textcolor{darkred}{+25.26\%} \\
Inference Throughput (samples/s) & 1.83 & 1.33 & \textcolor{darkred}{--27.32\%} \\
Parameters (M)        & 0.80  & 3.48  & \textcolor{darkred}{+335\%} \\
\bottomrule
\end{tabular}
\end{table}

In Table~\ref{tab:fft_variants_comparison}, we compare the performance and parameter counts of the CLAM~\cite{lu2021data} baseline against ZoomMIL~\cite{thandiackal2022differentiable}, a multiscale MIL approach, and FFT-MIL-mini, a reduced variant of our method with the maximum channel dimension of the CNN in the FFT-Block decreased from $32$ to $6$. FFT-MIL-mini retains performance comparable to the full FFT-MIL while increasing the parameters of CLAM~\cite{lu2021data} by only $25\%$. By contrast, ZoomMIL~\cite{thandiackal2022differentiable} underperforms relative to the other methods, which we attribute to limited robustness, as it was not previously evaluated on BRACS~\cite{brancati2022bracs}. Moreover, ZoomMIL~\cite{thandiackal2022differentiable} introduces a $261\%$ increase in parameters over CLAM~\cite{lu2021data}, consistent with its use of two additional magnification levels, which substantially raises model complexity.

\begin{table}[H]
\centering
\caption{Performance and complexity comparison of CLAM~\cite{lu2021data}, FFT-MIL, FFT-MIL-mini, and ZoomMIL~\cite{thandiackal2022differentiable} on the BRACS~\cite{brancati2022bracs} dataset. Metrics reported are weighted-averaged F1 score (F1), Area Under the Curve (AUC), and number of model parameters (Params). $\Delta$ values denote relative change compared to the CLAM baseline. FFT-MIL-mini denotes a reduced FFT-Block configuration with fewer channels, while ZoomMIL~\cite{thandiackal2022differentiable} is a multi-scale MIL approach.}
\vspace{10pt}
\label{tab:fft_variants_comparison}
\scriptsize
\renewcommand{\arraystretch}{1.5}
\setlength{\tabcolsep}{6pt}
\begin{tabular}{@{}l c c c c c c@{}}
\toprule
\textbf{Model} & \textbf{F1} & \textbf{AUC} & \textbf{Params (M)} & \textbf{$\Delta$F1} & \textbf{$\Delta$AUC} & \textbf{$\Delta$Params} \\
\midrule
CLAM  & 0.487 & 0.768 & 0.80 & -- & -- & -- \\
FFT-MIL (Ours)      & 0.525 & 0.815 & 3.48 & \textcolor{darkgreen}{+7.87\%}  & \textcolor{darkgreen}{+6.11\%} & \textcolor{darkred}{+335\%} \\
FFT-MIL-mini (Ours)    & 0.523 & 0.827 & 1.00 & \textcolor{darkgreen}{+7.44\%}  & \textcolor{darkgreen}{+7.68\%} & \textcolor{darkred}{+25\%} \\
ZoomMIL             & 0.347 & 0.811 & 2.89 & \textcolor{darkred}{-28.72\%} & \textcolor{darkgreen}{+5.57\%} & \textcolor{darkred}{+261\%} \\
\bottomrule
\end{tabular}
\end{table}

These comparisons show that incorporating a global frequency representation into MIL methods requires only a minimal computational increase. They also highlight the efficiency of our approach relative to multiscale methods, which depend on substantially larger architectures to capture global dependencies.  

\subsection{Frequency-Only vs. Spatial-Only Performance}
\label{subsec:abl2}
We evaluate the FFT-Block as a standalone frequency-only model and compare its performance to the spatial-only CLAM~\cite{lu2021data} for WSI classification in Table~\ref{tab:clam_vs_fftblock}. When used alone, the FFT-Block consistently underperforms relative to spatial methods, underscoring the importance of fine-grained detail. Its performance on the IMP~\cite{gillette2020proteogenomic} dataset is particularly limited. However, when integrated with MIL approaches, the FFT-Block still improves overall performance compared to spatial-only baselines, as shown in Table~\ref{tab:results2}, highlighting the value of coarse-grained frequency information for fusion.

\begin{table}[H]
\centering
\caption{Comparison of spatial-only CLAM~\cite{lu2021data} and our proposed frequency-only FFT-Block on BRACS~\cite{brancati2022bracs}, LUAD~\cite{gillette2020proteogenomic}, and IMP~\cite{neto2024interpretable} with accuracy (ACC) and F1 score (F1). $\Delta$ACC denotes the accuracy difference of the FFT-Block relative to CLAM~\cite{lu2021data}. The lower performance of frequency-only models is attributed to the loss of fine-grained spatial details that are effectively captured by patch-based methods. However, as shown in Table~\ref{tab:results2}, combining frequency and spatial representations yields the best overall results, as frequency-domain features capture global contextual dependencies.}
\vspace{10pt}
\label{tab:clam_vs_fftblock}
\scriptsize
\renewcommand{\arraystretch}{1.5}
\setlength{\tabcolsep}{6pt}
\begin{tabular}{@{}l l c c c@{}}
\toprule
\textbf{METHOD} & \textbf{DATASET} & \textbf{ACC} & \textbf{F1} & \textbf{$\Delta$ACC} \\
\midrule
CLAM~\cite{lu2021data}    & BRACS~\cite{brancati2022bracs} & 54.72\% & 0.536  & --         \\
FFT-Block (Ours)                 & BRACS~\cite{brancati2022bracs} & 50.94\% & 0.482  & \textcolor{darkred}{$-$3.78\%}  \\
CLAM~\cite{lu2021data}    & LUAD~\cite{gillette2020proteogenomic}  & 95.50\% & 0.955 & --         \\
FFT-Block (Ours)                 & LUAD~\cite{gillette2020proteogenomic}  & 91.89\% & 0.919  & \textcolor{darkred}{$-$3.61\%}  \\
CLAM~\cite{lu2021data}    & IMP~\cite{neto2024interpretable}   & 92.77\% & 0.928 & --         \\
FFT-Block (Ours)                 & IMP~\cite{neto2024interpretable}   & 68.67\% & 0.683  & \textcolor{darkred}{$-$24.10\%} \\
\bottomrule
\end{tabular}
\end{table}

\subsection{Robustness to Class Imbalance}
\label{subsec:abl3}
In Table~\ref{tab:results2}, we repeat the experiments from Table~\ref{tab:results} using the weighted-averaged F1 score for evaluation. Unlike the macro average, which treats all classes equally, the weighted F1 score prioritizes performance on frequent classes and better reflects real-world deployment settings where class imbalance is common~\cite{mcdermott2024closer}. The results show that FFT-MIL consistently outperforms all baselines, demonstrating robustness to class imbalance and improving the average WSI classification accuracy by $2.76\%$.

\begin{table}[H]
\centering
\caption{Evaluation of all methods as implemented by CLAM~\cite{lu2021data}, ACMIL~\cite{zhang2024attention}, and DGR-MIL~\cite{zhu2024dgr} on BRACS~\cite{brancati2022bracs}, LUAD~\cite{gillette2020proteogenomic}, and IMP~\cite{neto2024interpretable}, with Accuracy (ACC) and weighted-averaged F1 score (F1). $\Delta$ACC denotes the change in accuracy achieved by integrating FFT-MIL into each baseline MIL method, including CLAM~\cite{lu2021data}, MIL~\cite{ilse2018attention}, ABMIL~\cite{ilse2018attention}, ACMIL~\cite{zhang2024attention}, IBMIL~\cite{lin2023interventional}, and ILRA~\cite{xiang2023exploring}, over the three datasets, BRACS~\cite{brancati2022bracs}, LUAD~\cite{gillette2020proteogenomic}, and IMP~\cite{neto2024interpretable}. Best results are marked in bold. Methods marked with “(Ours)” denote the integration of the proposed FFT-MIL framework into the corresponding baseline.}
\vspace{10pt}
\label{tab:results2}
\scriptsize
\renewcommand{\arraystretch}{1.5}
\setlength{\tabcolsep}{4pt}
\begin{tabular}{@{}lcccccccc@{}}
\toprule
\multirow{2}{*}{\textbf{Method}} &
\multicolumn{2}{c}{\textbf{BRACS~\cite{brancati2022bracs}}} &
\multicolumn{2}{c}{\textbf{LUAD~\cite{gillette2020proteogenomic}}} &
\multicolumn{2}{c}{\textbf{IMP~\cite{neto2024interpretable}}} &
\multirow{2}{*}{\textbf{$\Delta$ ACC}} \\
& \textbf{ACC} & \textbf{F1} & \textbf{ACC} & \textbf{F1} & \textbf{ACC} & \textbf{F1} & \\
\midrule
\multicolumn{8}{c}{\textbf{CLAM}~\cite{lu2021data}} \\
\midrule
CLAM~\cite{lu2021data}              & 54.72\% & 0.536 & 95.50\% & 0.955 & 92.77\% & 0.928 & --       \\
CLAM (Ours)         & \textbf{62.26\%} & \textbf{0.601} & \textbf{97.30\%} & \textbf{0.973} & \textbf{95.18\%} & \textbf{0.953} & \textcolor{darkgreen}{+3.92\%} \\
MIL~\cite{ilse2018attention}               & 49.06\% & 0.479 & 95.50\% & 0.955 & 85.54\% & 0.857 & --       \\
MIL (Ours)          & 50.94\% & 0.497 & 96.40\% & 0.964 & 91.57\% & 0.916 & \textcolor{darkgreen}{+2.94\%}  \\
\midrule
\multicolumn{8}{c}{\textbf{ACMIL}~\cite{zhang2024attention}} \\
\midrule
ABMIL~\cite{ilse2018attention}             & 44.23\% & 0.305 & 92.79\% & 0.929 & 87.95\% & 0.881 & --       \\
ABMIL (Ours)        & 46.67\% & 0.323 & 94.59\% & 0.946 & 93.98\% & 0.941 & \textcolor{darkgreen}{+3.42\%}  \\
ACMIL~\cite{zhang2024attention}             & 42.31\% & 0.303 & 95.50\% & 0.955 & 78.31\% & 0.747 & --       \\
ACMIL (Ours)        & 45.71\% & 0.308 & 95.50\% & 0.955 & 86.75\% & 0.857 & \textcolor{darkgreen}{\textbf{+3.95\%}} \\
IBMIL~\cite{lin2023interventional}             & 44.23\% & 0.305 & 92.79\% & 0.929 & 87.95\% & 0.881 & --       \\
IBMIL (Ours)        & 46.67\% & 0.323 & 94.59\% & 0.946 & 87.95\% & 0.882 & \textcolor{darkgreen}{+1.41\%}  \\
\midrule
\multicolumn{8}{c}{\textbf{DGR-MIL}~\cite{zhu2024dgr}} \\
\midrule
ABMIL~\cite{ilse2018attention}               & 58.49\% & 0.511 & 94.59\% & 0.946 & 91.57\% & 0.916 & --       \\
ABMIL (Ours)        & 60.38\% & 0.561 & \textbf{97.30\%} & \textbf{0.973} & 93.98\% & 0.941 & \textcolor{darkgreen}{+2.34\%}  \\
ILRA~\cite{xiang2023exploring}              & 56.60\% & 0.539 & 94.59\% & 0.946 & 93.98\% & 0.941 & --       \\
ILRA (Ours)         & 58.49\% & 0.537 & 95.50\% & 0.955 & \textbf{95.18\%} & 0.952 & \textcolor{darkgreen}{+1.33\%}  \\
\bottomrule
\end{tabular}
\end{table}

\section{Conclusion}
In summary, this work introduces Fourier Transform Multiple Instance Learning (FFT-MIL), a framework that augments existing MIL methods with a compact frequency-domain representation to address the challenge of modeling global context in whole slide images. By extracting low-frequency crops and processing them through the proposed FFT-Block, FFT-MIL provides efficient and complementary global features that can be seamlessly integrated with diverse MIL architectures. Extensive experiments across three public datasets and six state-of-the-art MIL methods demonstrate that incorporating frequency-domain information consistently improves classification performance while incurring only modest computational cost. These findings establish FFT-MIL as a practical and generalizable approach for enhancing WSI analysis, highlighting the potential of frequency-domain learning to advance computational pathology beyond the limitations of purely spatial models.

\subsection*{Disclosures}
The authors declare that there are no financial interests, commercial affiliations, or other potential conflicts of interest that could have influenced the objectivity of this research or the writing of this paper.

\subsection*{Code, Data, and Materials Availability}
The code developed for FFT-MIL will be made publicly available upon acceptance. Experiments were conducted using publicly available datasets, including BRACS~\cite{brancati2022bracs}, LUAD~\cite{gillette2020proteogenomic}, and IMP~\cite{neto2024interpretable}. Additionally, existing open-source codebases were used to replicate and extend prior methods, including CLAM~\cite{lu2021data} (\linkable{github.com/mahmoodlab/CLAM}), ACMIL~\cite{zhang2024attention} (\linkable{github.com/dazhangyu123/ACMIL}), and DGR-MIL~\cite{zhu2024dgr} (\linkable{github.com/ChongQingNoSubway/DGR-MIL}).

\subsection*{Acknowledgments}
This research was supported by the University of Central Florida.

\bibliography{article}   
\bibliographystyle{spiejour}   
\listoffigures
\listoftables

\end{spacing}
\end{document}